\documentclass[manuscript,screen,nonacm]{acmart}%
\pdfoutput=1
\setcopyright{none}%
\usepackage{algorithm}
\usepackage{algorithmic}
\usepackage{graphicx}
\usepackage{textcomp}
\usepackage{xcolor}
\usepackage{pdflscape}
\usepackage{afterpage}

\newtheorem{definition}{Definition}
\newtheorem{proposition}{Proposition}

\usepackage{booktabs}
\usepackage{multirow}
\usepackage{multicol}

\usepackage{bbm}
\usepackage{tabularx}
\usepackage{hyperref} 
\usepackage[inline]{enumitem}

\usepackage{caption}
\usepackage{subcaption}

\newcommand{\sourcecode}{\url{https://github.com/xuanxuanxuan-git/sequential_cfe}
}

\graphicspath{{./pdfs/}}

\usepackage{fontawesome}

\usepackage[framemethod=tikz]{mdframed}
\definecolor{mycolor}{rgb}{0.122, 0.435, 0.698}
\definecolor{myblue}{rgb}{0.122, 0.435, 0.698}
\definecolor{mygreen}{rgb}{0.125, 0.525, 0.220}
\definecolor{myyellow}{rgb}{0.588, 0.439, 0.000}
\definecolor{myviolet}{rgb}{0.71764706, 0.40784314, 0.63529412}
\definecolor{myred}{rgb}{0.647, 0.114, 0.165}
\newmdenv[innerlinewidth=0.5pt,roundcorner=4pt,innerleftmargin=6pt,
          innerrightmargin=6pt,innertopmargin=6pt,innerbottommargin=6pt,
          linecolor=mycolor,backgroundcolor=mycolor!25!white]{mybox}
\newmdenv[innerlinewidth=0.5pt,roundcorner=4pt,innerleftmargin=6pt,
          innerrightmargin=6pt,innertopmargin=6pt,innerbottommargin=6pt,
          linecolor=myblue,backgroundcolor=myblue!25!white]{mybluebox}
\newmdenv[innerlinewidth=0.5pt,roundcorner=4pt,innerleftmargin=6pt,
          innerrightmargin=6pt,innertopmargin=6pt,innerbottommargin=6pt,
          linecolor=mygreen,backgroundcolor=mygreen!25!white]{mygreenbox}
\newmdenv[innerlinewidth=0.5pt,roundcorner=4pt,innerleftmargin=6pt,
          innerrightmargin=6pt,innertopmargin=6pt,innerbottommargin=6pt,
          linecolor=myyellow,backgroundcolor=myyellow!25!white]{myyellowbox}
\newmdenv[innerlinewidth=0.5pt,roundcorner=4pt,innerleftmargin=6pt,
          innerrightmargin=6pt,innertopmargin=6pt,innerbottommargin=6pt,
          linecolor=myred,backgroundcolor=myred!25!white]{myredbox}
\newmdenv[innerlinewidth=0.5pt,roundcorner=4pt,innerleftmargin=6pt,
          innerrightmargin=6pt,innertopmargin=6pt,innerbottommargin=6pt,
          linecolor=myviolet,backgroundcolor=myviolet!25!white]{myvioletbox}

\begin{document}

\title[Perfect Counterfactuals in Imperfect Worlds]{Perfect Counterfactuals in Imperfect Worlds: Modelling Noisy Implementation of Actions in Sequential Algorithmic Recourse}

\author{Yueqing Xuan}
\authornote{Corresponding author.}
\email{yueqing.xuan@student.rmit.edu.au}
\orcid{0000-0002-9365-8949}
\affiliation{%
  \institution{ARC Centre of Excellence for Automated Decision-Making and Society, School of Computing Technologies, RMIT University}
  \city{Melbourne}
  \country{Australia}
}

\author{Kacper Sokol}
\email{kacper.sokol@inf.ethz.ch}
\orcid{0000-0002-9869-5896}
\affiliation{
  \institution{Department of Computer Science, ETH Zurich}
  \city{Zurich}
  \country{Switzerland}
}

\author{Mark Sanderson}
\email{mark.sanderson@rmit.edu.au}
\orcid{0000-0003-0487-9609}
\author{Jeffrey Chan}
\email{jeffrey.chan@rmit.edu.au}
\orcid{0000-0002-7865-072X}
\affiliation{%
  \institution{ARC Centre of Excellence for Automated Decision-Making and Society, School of Computing Technologies, RMIT University}
  \city{Melbourne}
  \country{Australia}
}

\begin{abstract}
Algorithmic recourse suggests actions to individuals who have been adversely affected by automated decision-making, helping them to achieve the desired outcome. Knowing the recourse, however, does not guarantee that users can implement it perfectly, either due to environmental variability or personal choices. Recourse generation should thus anticipate its sub-optimal or noisy implementation. %
While several approaches construct recourse that is robust to small perturbations -- e.g., arising due to its noisy implementation -- they assume that the entire recourse is implemented in a single step, thus model the noise as one-off and uniform. %
But these assumptions %
are unrealistic since recourse often entails multiple sequential steps, which makes it harder to implement and subject to increasing noise. %
In this work, we consider recourse under plausible noise that adheres to the local data geometry and accumulates at every step of the way. %
We frame this problem as a Markov Decision Process and demonstrate that such a distribution of plausible noise satisfies the Markov property. %
We then propose the RObust SEquential (ROSE) recourse generator for tabular data; our method produces a series of steps leading to the desired outcome even when they are implemented imperfectly. %
Given plausible modelling of sub-optimal human actions and greater recourse robustness to accumulated uncertainty, 
ROSE provides users with a high chance of success while maintaining low recourse cost. %
Empirical evaluation shows that our algorithm effectively navigates the inherent trade-off between recourse robustness and cost while ensuring its sparsity and computational efficiency. %
\end{abstract}


\maketitle

\begin{mybluebox}
\noindent\faGithub\hspace{.2cm}\textbf{Source Code}\quad%
\sourcecode%
\end{mybluebox}

\vspace{.33em}%
\begin{myvioletbox}
\noindent\faFileTextO\hspace{.2cm}\textbf{Published in}\quad%
ECML-PKDD 2025: Journal Track %
(Springer Machine Learning -- \href{https://doi.org/10.1007/s10994-025-06821-1}{10.1007/s10994-025-06821-1})
\end{myvioletbox}

\section{Introduction}

Automated decision-making has become ubiquitous in high stakes domains such as finance and healthcare. Consequently, it is important to empower the affected individuals by offering them counterfactual explanations (CE) in the form of algorithmic recourse to help them achieve the desired outcomes~\citep{wachter2017counterfactual}. Prior work on CEs shows explainees what features to alter, and by how much, to change the model's prediction~\citep{joshi2019towards,ustun2019actionable}. For example, when a person's loan application is rejected by a bank's model, they should be provided with a CE that informs them to increase their income by \$1,000 and reduce their consumer debt by \$5,000 to receive the loan. %

However, simply highlighting feature-wise differences between factual and counterfactual instances suggests that these changes can be achieved instantly and simultaneously in a single step~\cite{verma2020counterfactual}. In practice, users will not necessarily be able to make all of the suggested changes instantaneously; instead they may need to make incremental changes through a sequence of discrete actions~\cite{barocas2020hidden}. In the previous example, a realistic guidance to a user might be to reduce debt first and then increase income. This is known as 
(prospective) sequential recourse that generates a sequence of discrete actions leading to a counterfactual outcome~\cite{ramakrishnan2020synthesizing,kanamori2021ordered,verma2022amortized}. 

Current methods generate counterfactuals under the assumption that users will implement them faithfully; however, in the real world recourse is often subject to noisy implementation~\cite{pawelczyk2023probabilistic}. 
Prior research has noted that the implementation of recourse involving multiple actions is often affected by noise~\cite{bjorkegren2020manipulation}, and that humans tend to follow sequential instructions imperfectly~\cite{degrandpre1991effects}. %
To complicate matters further, individuals might lack precise control over the value of a feature when taking an action~\cite{barocas2020hidden}, or feature values might change unexpectedly due to unforeseen circumstances~\cite{virgolin2023robustness}. %
Consequently, uncertainty could arise while executing actions, causing feature values to diverge from what a user is expecting after a discrete step. For example, a user who planned to increase their income by \$1,000 may eventually end up with a factual rise of \$990 due to inflation. %
In this case, users should ideally still end up in the desired position once they have completed the original recourse, i.e., counterfactual generation
methods should be robust to users' genuine but possibly sub-optimal intentions. %

\begin{figure}[t]
    \centering
    \begin{subfigure}{0.4\columnwidth}
    \centering 
        \includegraphics[height=3.7cm]{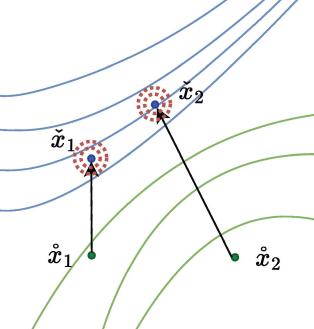}
        \caption{Identical Gaussian noise added to recourse of varying length.}%
        \label{fig:exp1}
    \end{subfigure}
    \hspace{1.5cm}
    \begin{subfigure}{0.4\columnwidth}
    \centering
        \includegraphics[height=3.7cm]{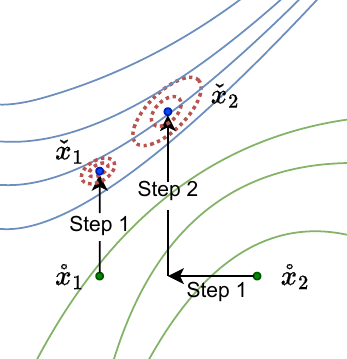}
        \caption{Plausible noise that adapts to the local data geometry and recourse length.}%
        \label{fig:exp2}
    \end{subfigure}
    \caption{Examples of robust recourse achieved with (\subref{fig:exp1})~an existing method and (\subref{fig:exp2})~our framework. The blue lines enclose the desirable region. %
    The contours indicate data density. Panel~(\subref{fig:exp1}) shows that prior work applies the same size of noise to recourse of varying length~\cite{guyomard2023generating,pawelczyk2023probabilistic}; plausibility of noise is also ignored. In Panel~(\subref{fig:exp2}), we consider that longer recourse is subject to larger uncertainty and we model plausible noise that follows the local data geometry. %
    }
\end{figure}

Current literature also agrees that longer recourse is harder to implement as it often entails more actions or the scale of each action is larger, which increases the complexity of its implementation. Intuitively, users are less likely to faithfully follow the recourse as the process gets longer, thus uncertainty in human actions -- i.e., noise -- can accumulate along the recourse-seeking process.
However, existing recourse generators assume that simple recourse is equally uncertain to complex recourse and make the former robust to unnecessarily large noise at the expense of its size. %
On the other hand, complex recourse is only robust to smaller uncertainty, so users are still likely to fail (see Figure~\ref{fig:exp1}). To overcome this shortcoming %
we address robust recourse in a more realistic setting where each recourse consists of multiple discrete actions of varied size, %
and uncertainty is associated with each action, accumulating as the process gets longer (see Figure~\ref{fig:exp2}). 
Since the magnitude of noise that we model is adaptive to the recourse length, shorter recourse will not be penalised by additional cost (increased length) to address unrealistically large uncertainty. %

When infusing recourse with noise to model real-world noisy recourse implementation, the distribution of noise should be realistic and supported by sufficient data. To this end, we introduce plausible noise distribution and demonstrate the Markov property of plausible noise associated with recourse steps. We further model the robust recourse generation problem as a Markov Decision Process (MDP). %
Our work complements the studies by %
\citet{guyomard2023generating} and \citet{pawelczyk2023probabilistic}, which ignore plausibility or increasing uncertainty when modelling noisy action implementation.
It is also complementary to the work by \citet{dominguez2022adversarial}, which models causal noise under the explicit knowledge of causal models. 
Our noise formulation can be easily generalised without the knowledge of causal models and simultaneously captures plausibility.

Our contribution is threefold.
\begin{enumerate}
    \item 
In Section~\ref{sec:plau-noise} we model plausible noise to represent noisy implementation of human actions. %
\item
Section~\ref{sec:path-noise} formulates the distribution of accumulated plausible noise and shows 
its Markov property; then we use an MDP to model the accumulation of plausible noise along the recourse-seeking process. %
\item
In Section~\ref{sec:mdp} we propose the RObust SEquential (ROSE) recourse generator designed for high-dimensional tabular datasets; %
it offers users sequential recourse that is robust to noisy action implementation. %
\end{enumerate}
As a result, 
our robust recourse is more realistic than instantaneous counterfactuals, and grants users higher chances of success even if they cannot implement every action precisely.
Our plausible modelling of sub-optimal human actions also allows robust recourse to remain of low cost, i.e., requiring small feature changes.
An extensive evaluation with three real-world datasets and seven baselines demonstrates superior performance of our method in terms of robustness and feature sparsity while effectively managing the trade-off between robustness and recourse cost.

\section{Related Work}\label{sec:related-work}

\subsection{Sequential Counterfactual Explanations}
Existing CE methods suggest changes to feature values that lead to a desired outcome. Common objectives include proximity between factual and counterfactual instances~\cite{russell2019efficient,ustun2019actionable,karimi2020model}, number of changed features (sparsity)~\cite{dandl2020multi,mothilal2020explaining,van2021interpretable}, and feasibility of counterfactual instances through data density or closeness to existing instances~\cite{joshi2019towards,downs2020cruds,pawelczyk2020learning,van2021interpretable,bobek2023local}. However, these methods only return the counterfactual instances, which reflects their assumption that recourse is an instantaneous, one-step process~\cite{verma2020counterfactual}. In reality, changes do not happen instantly but through a sequence of discrete steps~\cite{verma2020counterfactual,Naumann2021}. Recent papers have proposed sequential counterfactual generation. 
For example, FACE provides a sequence of existing data instances as a path to the target counterfactual~\cite{poyiadzi2020face}, %
and different optimisation techniques, such as MDP, exist~\cite{ramakrishnan2020synthesizing,kanamori2021ordered,Naumann2021,verma2022amortized}.
Despite recent interest in producing sequential actions to support recourse, little attention has been paid to the possibility of its noisy implementation, which may easily invalidate recourse. %

The aforementioned work that produces instantaneous counterfactuals is commonly based on supervised learning. Reinforcement learning has recently gained popularity in CE generation where sequential recourse is an integral part. For example, \citet{tsirtsis2024finding} focus on time series data and explore ``if some historical actions had changed, how the current situation would have been like''. In other words, they address retrospective CEs in sequential settings. On the other hand, our work focuses on prospective CEs that inform users what to do in the next steps to get a better outcome in the future. To the best of our knowledge, our paper is the first to model sequential CEs prospectively under serial uncertainty. %

\subsection{Robustness of Counterfactual Explanations}
Current literature recognises various events to which CEs should be robust. %
Examples include robustness to shifts of prediction models~\cite{upadhyay2021towards}, data distribution shifts~\cite{rawal2020algorithmic}, model multiplicity~\cite{pawelczyk2020counterfactual,jiang2024recourse}, and (gradual) model re-training~\cite{ferrario2022robustness}.
In addition, \citet{sharma2022faster} explored whether CE generators give consistent outputs when users check for updated CEs along their original recourse.
\citet{raman2023bayesian} defined robustness as CEs residing in dense data regions -- a property that is commonly referred to as \emph{plausibility}~\cite{wielopolski2024probabilistically}. 

Our work is closely related to another notion of robustness that deals with the perturbation of feature values. Such perturbation applied to the explained instance has been shown to manipulate or invalidate the corresponding CEs~\cite{slack2021counterfactual,pawelczyk2023probabilistic}. %
In this space, \citet{dominguez2022adversarial} proposed a method to make counterfactuals robust to (causal) uncertainty in feature input; \citet{guyomard2023generating} and \citet{pawelczyk2023probabilistic} proposed to handle perturbations applied to the counterfactual instance, which simulates noisy implementation of recourse; and %
\citet{virgolin2023robustness} studied robustness to perturbations of features that are altered and kept unchanged in the recourse. %
However, these methods assume that the implementation of counterfactuals is a one-shot process, thus they only add a constant amount of noise once to every counterfactual despite their differences in length and sparsity.
This means that simpler (shorter) recourse is made robust to unnecessarily large perturbations whereas complex recourse becomes invalidated as uncertainty accumulates. %
With recent work paying more attention to providing recourse as a sequence of actions, %
it is important to address the robustness of \emph{sequential} recourse to \emph{accumulated} noise. 

Additionally, literature on robust counterfactuals uses inconsistent formulation of adversarial perturbations to individual data instances. \citet{pawelczyk2020counterfactual} and \citet{guyomard2023generating} draw noise once per instance from a Gaussian probability distribution. However, they assume the same magnitude of uncertainty for different recourse options regardless of how difficult they are to complete.
\citet{dominguez2022adversarial} take into account the linear causal relationship between features 
by leveraging the explicit knowledge of causal models; however, the occurrence of noisy instances from a perturbation set is treated as equally possible. 
\citet{virgolin2023robustness} manually define the perturbation ranges for each feature based on domain knowledge, and the perturbation set contains all possible combinations of perturbed feature values. Similarly, they assume the perturbation points within the set are equally likely to occur.
Our work incorporates plausibility and accumulated uncertainty into noise formulation, which models human's sub-optimal recourse implementation more realistically. This approach does not rely on explicit knowledge of causal models or domain information, thus has better generalisability. We further provide a graphical summary of various types of perturbations in Appendix~\ref{app:comparison} to facilitate better comparison.

It is also important to distinguish recourse that is robust against plausible noise and plausible recourse. Our work focuses on plausible uncertainty in recourse implementation, whereas other studies %
consider plausibility in recourse generation, assuming perfect recourse implementation~\cite{bobek2023local,raman2023bayesian,wielopolski2024probabilistically}. For example, PPCEF algorithm generates probabilistically plausible counterfactual explanations without assuming noise in actual recourse implementation~\cite{wielopolski2024probabilistically}.
Since plausibility can be investigated under different problem definitions, our paper focuses specifically on approaches that deal with imperfect recourse implementation. %

\begin{figure}[t]
    \centering
    \begin{subfigure}{0.45\columnwidth}
    \centering
        \includegraphics[height=4.5cm]{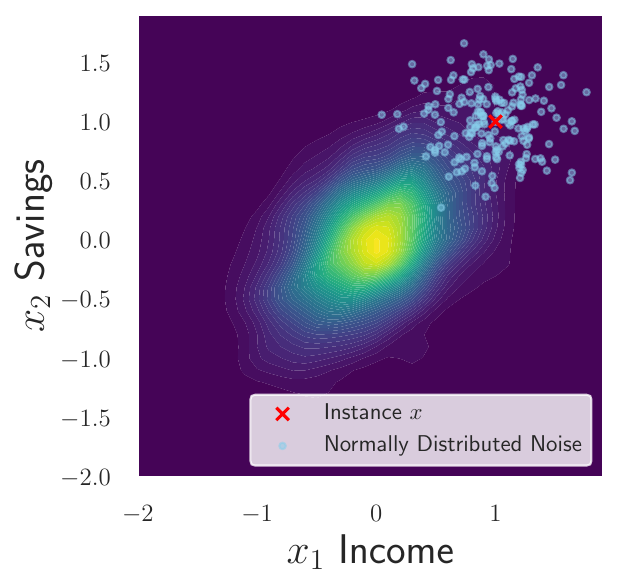}
        \caption{Normally distributed noise.}
        \label{fig:noise-1}
    \end{subfigure}
    \begin{subfigure}{0.45\columnwidth}
    \centering
        \includegraphics[height=4.5cm]{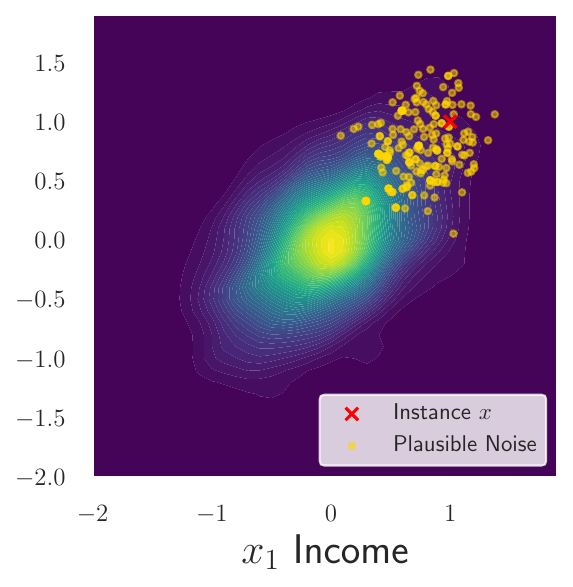}
        \caption{Plausible noise.}
        \label{fig:noise-3}
    \end{subfigure}
    \caption{
    Two approaches to sample random noise around an instance $x$ (marked by a red cross). %
    This synthetic toy dataset has two positively correlated features. The contours indicate the density of the data distribution estimated via kernel density estimation.     
    Current literature assumes (\subref{fig:noise-1}) normally distributed noise around the selected point, but this distribution fails to account for plausibility.
    For example, instances with higher income and savings than the chosen point are less plausible despite being close, i.e., there is insufficient data support. %
    Our approach models (\subref{fig:noise-3}) plausible noise by considering both its distance to the selected point and local data distribution.}
    \label{fig:resampled_noise}
\end{figure}

\section{Noisy Recourse Implementation\label{sec:defs}}
In this section, we first define the generic formulation of recourse and MDPs (Section~\ref{sec:defs:gen}). 
We then propose novel ways to generate one-off (Section~\ref{sec:plau-noise}) and accumulated (Section~\ref{sec:path-noise}) plausible noise to capture imperfect recourse implementation.

\subsection{General Formulation\label{sec:defs:gen}}

\paragraph{Notation}
Let $f: \mathcal{X} \rightarrow \mathcal{Y}$ denote a classifier that maps features $x \in \mathcal{X} \subseteq \mathbb{R}^d$ to labels $\mathcal{Y}$. Let $\mathcal{Y} = \{0, 1\}$ where 0 and 1 respectively denote an unfavourable outcome (e.g., loan denied) and a favourable outcome (e.g., loan approved).
Let $\mathring{x} \in \mathcal{X}$ denote the factual instance for which we generate recourse, i.e., $f(\mathring{x})=0$. We denote $h:\mathcal{X} \rightarrow \mathcal{X}$ a recourse generator that returns a counterfactual instance $\check{x}$ such that $f(\check{x})=1$. %
Using a pre-defined distance function $d:\mathbb{R}\rightarrow\mathbb{R}_+$, we consider $\delta = d(\mathring{x}, \check{x})$ to be the recourse length, i.e., its cost. 

\paragraph{Markov Decision Process}
In this paper, we frame the recourse-finding problem as an MDP (see Section~\ref{sec:path-noise} for the justification of this choice). %
In an MDP problem, a user acting towards recourse is called an \emph{agent} who corresponds to a $d$-dimensional data point $x$. The combination of possible values for all features in $\mathcal{X}$ forms the \emph{state space} $\mathcal{S}$ for the MDP. Upon taking a specific action $a$, an agent can move from one state $s$ to another state $s^{\prime}$. These actions constitute the \emph{action space} $\mathcal{A}$ for the MDP.
The third component of the MDP is the \emph{transition function}: $s \times a \rightarrow s^{\prime}$. 
The final component of the MDP is the \emph{reward function}. It associates each action with a reward, where reaching desirable states generates a positive reward. 

\paragraph{Connecting CEs to MDPs}
When modelling the CE generation for a user through an MDP, we assume that the user (agent) starts 
from the initial state $s_0\equiv \mathring{x}$, and takes a sequence of $k$ (ordered) actions $A=\{a_1,\ldots, a_k \; | \; s_{i-1} \times a_i \rightarrow s_i\}$ to get to the counterfactual state $s_k\equiv\check{x}$. 
Our goal is to find a policy $\pi: \mathcal{S} \rightarrow A$ that, given a starting state $s_0 \in \mathcal{S}$ (a factual data point), returns the best sequence of actions $A$ to reach a counterfactual state. In Section~\ref{sec:mdp}, we will detail the formulation of each component in the MDP to generate robust recourse.

\subsection{Plausible Noise}\label{sec:plau-noise}

Current literature on robust recourse often models perturbations by adding small normally distributed noise $\epsilon \sim \mathcal{N}(0, \sigma^2 \cdot \textbf{I})$ to the counterfactual instance $\check{x}$. This implies that instances close to $\check{x}$ are also expected to be of the desirable class. 
Given that recourse takes place in the real world, points representing the noisy human implementation of $\check{x}$ should also be realistic in the sense that they lie within the data manifold and follow plausible data distribution. Consider the examples shown in Figure~\ref{fig:resampled_noise} where savings and income are positively correlated. Given the data manifold geometry, we expect noisy instances around $x$ with both higher income and higher savings to be less likely than those with lower income and lower %
savings because the former situation lacks sufficient data support. Thus the plausible noise distribution in Figure~\ref{fig:noise-3} is more informative and better reflects realistic human implementation of recourse.
Definition~\ref{def:plausible-noise} formalises our notion of plausible noise. %

\begin{definition}[Plausible Noise]\label{def:plausible-noise}
The plausible noise surrounding an instance $x$ can be modelled by some noise $\epsilon$
characterised by the probability distribution $p_{x}$ where
\begin{equation}\label{eq:plau-dist}
    \epsilon = x^{\prime} - x %
    \quad \text{s.t.} \quad x^{\prime} \sim p_{x} = c \cdot \mathcal{N}(x, \sigma^2\cdot \mathbf{I}) \cdot K(\mathcal{X})~\text{.}
\end{equation}
$\mathcal{N}(x, \sigma^2\cdot \mathbf{I})$ is the Gaussian probability distribution with mean vector $x$ and covariance matrix $\sigma^2\cdot\mathbf{I}$;
$K$ is an estimate of the probability density function based on $\mathcal{X}$, which reflects the density of the underlying data distribution; and $c$ is the normalisation constant to ensure $p_x(x^\prime)$ integrates to $1$.
\end{definition}

Intuitively, a small perturbation $\epsilon$ to $x$ results in plausibly similar instances $x^{\prime}$ located close to $x$. %
Specifically, the closer a random sample $x^{\prime}$ is to $x$, and the more likely it obeys the local data distribution geometry around $x$, the more likely it is to be drawn. %
Unlike normally distributed noise~\cite{pawelczyk2023probabilistic}, our formulation captures noise under a more realistic setting.
While noise modelled by a linear Structural Causal Model (SCM) may better account for plausibility~\cite{dominguez2022adversarial}, knowing the exact SCM is often infeasible~\cite{verma2022amortized}; similarly, defining uncertainty range manually for every feature based on domain knowledge is impractical~\cite{virgolin2023robustness}. Therefore our noise formulation can operate in a more general setting where the underlying causal model remains unknown.

\paragraph{Evaluating Plausible Noise} 

When evaluating the robustness of recourse to plausible noise, we perturb $\check{x}$ by $\epsilon$. We call this type of perturbation ``one-off noise'' because it is applied only once to the entire recourse. This set-up is prevalent in existing work on robust recourse given its assumption that recourse is a one-step process itself. 
Following this setting, the magnitude of the plausible noise $\sigma$ is neither affected by the size of recommended changes $\delta$ nor by how a user progresses towards $\check{x}$ (i.e., $A$); the distribution is realistic in so far as it should lie within the data manifold. 
We revise this assumption in Section~\ref{sec:path-noise}.

To find a counterfactual instance that is robust to one-off plausible noise, we expect $\check{x} + \epsilon$ to be classified with the counterfactual class. Such robustness can be measured with Invalidation Rate (IR), which is defined by \citet{pawelczyk2023probabilistic} as
\begin{equation}\label{eq:ir-eq}
    \text{IR}(\check{x}) = \mathbb{E}_{\epsilon}[f(\check{x}) - f(\check{x} + \epsilon)]\text{,}   
\end{equation}
where the expectation is taken with respect to a random variable $\epsilon$ controlled by the probability distribution $p_{\check{x}}$ given in Equation~\ref{eq:plau-dist}. We further derive a closed-form expression for IR and its approximation in Proposition~\ref{prop:ir}. 

\begin{proposition}\label{prop:ir}
    If $K(\mathcal{X})$ is a smooth estimate of the probability density over $\mathcal{X}$ -- e.g., kernel density estimation -- such that $p_{\check{x}}$ is continuous, we can compute IR($\check{x}$) as
    \begin{equation}\label{eq:plau-ir-inte}
    \begin{aligned}
        \text{IR}(\check{x}) = & 1- \int_{-\infty}^{\infty} 1 \cdot \mathbbm{1}\left(f(x)=1\right)\cdot p_{\check{x}}(x)dx 
        - \int_{-\infty}^{\infty} 0 \cdot \mathbbm{1}\left(f(x)=0\right)\cdot p_{\check{x}}(x)dx \\
         = & 1 - \int_{-\infty}^{\infty}f(x)\cdot p_{\check{x}}(x)dx\text{.}
    \end{aligned}
    \end{equation}
    If $p_{\check{x}}$ is discrete, we can approximate the expected IR through Monte Carlo Sampling as
    \begin{equation}\label{eq:ir-discrete}
        \widetilde{\text{IR}}(\check{x}) = 1 - \frac{1}{N}\sum_{i=1}^{N}f(x_i)\text{,}
    \end{equation}
    where $x_i$ is sampled from the $p_{\check{x}}$ distribution and $N$ is the number of samples.
\end{proposition}

\begin{figure*}[t]
    \centering
    \begin{subfigure}{0.31\linewidth}
        \includegraphics[height=.9\linewidth]{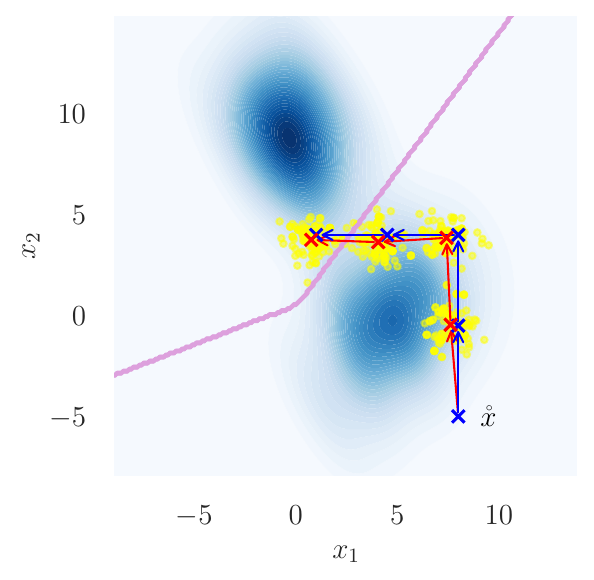}
        \caption{Increase $x_2$ first, then decrease $x_1$.}
        \label{fig:different-path-a}
    \end{subfigure}
    \hfill
    \begin{subfigure}{0.31\linewidth}
        \includegraphics[height=.9\linewidth]{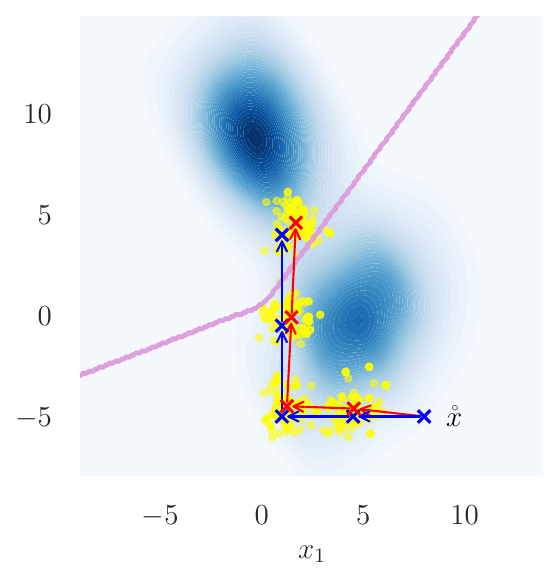}
        \caption{Decrease $x_1$ first, then increase $x_2$.}
        \label{fig:different-path-b}
    \end{subfigure}
    \hfill
    \begin{subfigure}{0.31\linewidth}
        \includegraphics[height=.9\linewidth]{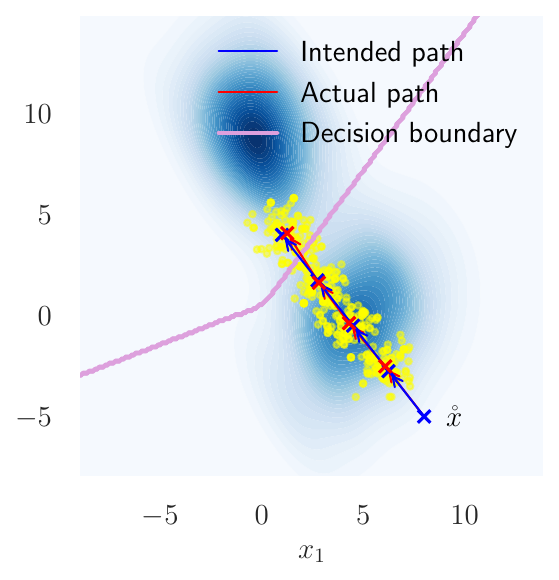}
        \caption{Change $x_1$ and $x_2$ at the same time.}
        \label{fig:different-path-c}
    \end{subfigure}
    \caption{Recourse examples for a factual instance $\mathring{x}$ targeting the top-left desirable class patch. 
    Actions can modify (\subref{fig:different-path-a} \& \subref{fig:different-path-b}) a single feature or (\subref{fig:different-path-c}) multiple features at each step, but it is desirable for an action to change the smallest possible set of features. %
    Here, recourse is split into four discrete actions, which reflects how it is likely to be implemented in reality. The blue path shows the intended recourse trajectory when every action is implemented faithfully. The red path demonstrates the actual trajectory if every action is implemented with noise. %
    For each action, we randomly sample 100 points from its plausible noise distribution and compute their mean, which yields a noise-aware step in which the follow-on actions are anchored.
    For demonstration purposes, we use the mean of plausible noise as the resulting state for each action, but in actual experiments we allow the transition state to be sampled randomly from the noise distribution. %
    The distribution of plausible noise associated with each action differs depending on the order of previous actions, which cumulatively affect
    the recourse trajectory. Therefore, the order of actions should be considered when dealing with the robustness of recourse to noisy implementation of actions. %
    }
    \label{fig:different_path}
\end{figure*}

\subsection{Accumulated Plausible Noise}\label{sec:path-noise}

When adding one-off noise, recourse is assumed to consist of a single step and the magnitude of noise is length-agnostic. %
Recall that \citet{pawelczyk2023probabilistic} assume $\epsilon$ to follow normal distribution $\mathcal{N}(0, \sigma^2\mathbf{I})$ and set $\sigma$ to be the same for every recourse. 
Under this assumption, increasing the salary level by \$500 is associated with the same degree of uncertainty as rising it by \$50,000. Similarly, \citet{dominguez2022adversarial} and \citet{virgolin2023robustness} add perturbations of the same size %
to different recourse realisations regardless of the magnitude of their feature changes. %

Here, we argue that noise -- arising due to imperfect human implementation of recourse -- accumulates along its path. 
Recall that in real-world settings recourse involving sequential actions is often noisily implemented, and that humans tend to initially follow sequential instructions but fail to comply with them as time passes~\cite{degrandpre1991effects,hugtenburg2013definitions,bjorkegren2020manipulation}. Given the prevalence of such phenomena, we introduce a novel approach to algorithmically account for such user behaviour. Formally, we state that recourse implementation is affected by the problem where the degree of user adherence (or precise implementation) gradually decreases as the process gets longer.

In addition, recourse generation should consider the discreteness of actions and provide recourse as a series of actions~\cite{verma2020counterfactual}. As such, noise could arise when implementing every single action -- the more actions a user needs to take and the bigger the changes that each action entails, the more noise the entire process is likely to accumulate. For example, recourse that requires an increase of both salary and savings by \$50,000 would take longer to complete than recourse that only affects one of these features, thus unforeseen circumstances are more likely to arise in the former case.  
Similarly, if savings are to be increased by only \$500 rather than \$50,000, such recourse becomes easier to implement and less uncertainty is expected.
All such factors -- captured in Proposition~\ref{prop:acc-noise} -- impact the ability of a user to faithfully implement recourse, with the \emph{accumulated plausible noise} formalised in Definition~\ref{def:path-noise}.

\begin{proposition}[Accumulated Noise]\label{prop:acc-noise}
    Assuming that noise can arise during the implementation of every recourse step, this noise would accumulate at every step proportionally to the number of features being changed and the magnitude of these changes. %
\end{proposition}

\begin{definition}[Accumulated Plausible Noise]\label{def:path-noise}
    For recourse from $\mathring{x}$ to $\check{x}$ that requires a series of actions $A=\{a_1, \ldots, a_k\}$, where each action $a_i\in A$ is subject to noisy implementation, 
    the accumulated noise distribution is
    \begin{equation}\label{eq:path-noise-dist}
    \begin{aligned}
        & \mathcal{N}_{(\mathring{x}, A)}(\check{x}) = \sum_{i=1}^{k} 
        \epsilon_{s_i} + \check{x} \\
        \text{s.t.} \quad & \epsilon_{s_i} = s_i^{\prime} - s_i \qquad
            s_i^{\prime} \sim \mathcal{N}(s_i, \sigma_i^2\mathbf{I})\cdot K(\mathcal{X}) \\
            & s_i \gets s_{i-1}^{\prime} \times a_i \qquad
            \sigma_i^2 = \frac{||a_i||}{u} \times \sigma^2\text{,} %
    \end{aligned}
    \end{equation}
    where $u$ is the size unit of a standard action, $\sigma^2$ is the magnitude of noise associated with an action of size $u$, and $s_0 \equiv s_0^{\prime} \equiv \mathring{x}$.
\end{definition}

From Equation~\ref{eq:path-noise-dist} we can also see that the distribution of $s_i^{\prime}$ depends on $s_i$, where $s_i$ further depends on $s_{i-1}^{\prime}$ and $a_i$. Recursively, $s_{i-1}^{\prime}$ follows the probability distribution characterised by $s_{i-1}$, and the latter is influenced by $s_{i-2}^{\prime}$ and $a_{i-1}$.
At the end, the distribution of the accumulated plausible noise is determined by the locations of intermediate points $s_{1},\ldots, s_{k}$, which are ultimately controlled by the order of actions in $A$. Once $s_{i-1}^{\prime}$ and $a_i$ are known, $s_i$ can be set, then the distribution of $s_i^{\prime}$ is \emph{conditionally independent} of all the previous actions $a_{1},\ldots, a_{i-1}$ and past states $s_{1}^\prime,\ldots, s_{i-2}^{\prime}$. This shows that the noisy transition from one state to the next satisfies the Markov property, which is captured by Proposition~\ref{prop:acc-noise-dist}.

\begin{proposition}[Markov Property of Noisy State Transition]\label{prop:acc-noise-dist}
    If the noise associated with each recourse action follows the plausible noise distribution, the distribution of the accumulated plausible noise depends on the order and scale of said actions. Given $s_{i-1}^{\prime}$ and $a_i$, the subsequent $\epsilon_{s_i}$ and $s_{i}^{\prime}$ do not depend on the past. %
    In this case the noisy transition from $s_{i-1}^{\prime} \times a_i$ to $s_i^{\prime}$ satisfies the Markov property. 
\end{proposition}

We demonstrate sub-optimal recourse-seeking processes in Figure~\ref{fig:different_path}, which alter feature values in different order; these examples show the influence of action order on the distribution of accumulated plausible noise. Proofs of Propositions~\ref{prop:acc-noise} and~\ref{prop:acc-noise-dist} 
are given in Appendix~\ref{app:proof}.

\begin{algorithm}[t]
\caption{Noisy implementation of an action under plausible noise.}\label{alg:acc-def}
\begin{algorithmic}[1]
\REQUIRE current state CurrState, probability density function \texttt{K}($\mathcal{X}$), action $a$, noise variance unit $\sigma^2$, action size unit $u$.
\ENSURE NextStatePrime
\STATE \textbf{Function} \texttt{NoisyStep}(CurrState, \texttt{K}, $a, \sigma$)
    \begin{ALC@g}
        \STATE NextState $\gets$ CurrState $\times a$
        \STATE var $\gets \sigma^2 \times \frac{||a||}{u}$
        \STATE $B(x)$ $\sim \mathcal{N}(\text{NextState}, \text{var} \cdot \mathbf{I}) \cdot \texttt{K}(\mathcal{X})$  \COMMENT{sample plausible noise}
        \STATE NoiseSet $\gets$ $B(x)$
        \STATE NextStatePrime $\gets$ \texttt{RandomChoice}(NoiseSet)
        \STATE \textbf{Return} NextStatePrime
    \end{ALC@g}
\end{algorithmic}
\end{algorithm}

\begin{algorithm}[t]
\caption{Evaluate robustness to accumulated plausible noise.}\label{alg:acc-eval}
\begin{algorithmic}[1]
\REQUIRE number of iterations $M$, factual instance $\mathring{x}$, set of actions $A$, noise variance unit $\sigma^2$, classifier $f$, target class $\check{y}$.
\ENSURE $\text{IR}_{A}(\check{x})$.
\STATE count $\gets 0$
\FOR{$m=0,\ldots, M-1$}
    \STATE CurrState $\gets \mathring{x}$   
    \FOR[noisy implementation of $A$]{$a_i$ in $A$} 
        \STATE CurrState $\gets$ \texttt{NoisyStep}(CurrState, \texttt{K}, $a_i, \sigma$)
        \COMMENT{see Algorithm~\ref{alg:acc-def}}
    \ENDFOR
    \IF{$f(\text{CurrState}) == \check{y}$}
    \STATE count $\gets \text{count} + 1$
    \ENDIF
\ENDFOR
\STATE  $\text{IR}_{A}(\check{x}) \gets \frac{\text{count}}{M}$
\end{algorithmic}
\end{algorithm}

\paragraph{Evaluating Accumulated Noise}

For a known action list $A$ guiding the recourse from $\mathring{x}$ to $\check{x}$, the Invalidation Rate is %
\begin{equation}\label{eq:path-ir-inte}
    \text{IR}_{A}(\check{x}) = 1- 
        \int_{-\infty}^{\infty} f(x)\cdot p_x\left(\mathcal{N}_{(\mathring{x}, A)}(\check{x})\right) d(x) \text{,}
\end{equation}
where the $p_x$ function outputs the likelihood of $x$ being drawn from the accumulated noise distribution. In a discrete setting, $p_x$ is computed empirically via Monte Carlo Sampling. Note that this cannot be done by sampling random points from $p_{\check{x}}$ (or $p_{s_k}$) since $p_{s_k}$ is non-deterministic unless the location of all of its predecessors $s_{1}^\prime,\ldots, s_{k-1}^{\prime}$ are fixed. 
To overcome this challenge we can repeat the complete implementation of $A$ multiple times, where each time $a_{1},\ldots, a_{k}$ is infused with randomly sampled plausible noise. Algorithm~\ref{alg:acc-def} shows the process of adding single plausible noise to an action, and Algorithm~\ref{alg:acc-eval} demonstrates the procedure for accumulating plausible noise and its evaluation.

\begin{algorithm}[t]
\caption{MDP specification of robust algorithmic recourse.}\label{alg:rose}
\begin{algorithmic}[1]
\REQUIRE probability density function \texttt{K}($\mathcal{X}$), factual instance $\mathring{x}\equiv s_{0}$, target class $\check{y}$, classifier $f$, invalidation rate function \texttt{IR}, noise variance unit $\sigma^2$, robustness threshold $\tau$, reward discount factor $\gamma$, a large constant reward ConstantR, Boolean for enabling accumulated noise \texttt{AccumulatedNoise}, continuous domains $\mathbb{R}^{|\text{Num}|}$, discrete domains $\mathbb{Z}^{|\text{Cat}|}$, action space $\mathcal{A}$.
\ENSURE MDP.
\STATE State space $\mathcal{S}\subseteq \mathbb{R}^{|\text{Num}|} \times \mathbb{Z}^{|\text{Cat}|}$
\STATE Actions $A(s_{0}) \subseteq \mathcal{A}$ for $s_{0}$
\STATE $a \in A(s_{0})$ \COMMENT{a single action}
\STATE \textbf{Function} \texttt{Transition} (CurrState, \texttt{K}, $a, \sigma$)
\begin{ALC@g}
    \IF[noisy step]{\texttt{AccumulatedNoise}}
        \STATE NextState $\gets$ \texttt{NoisyStep}(CurrState, \texttt{K}, $a, \sigma$)
        \COMMENT{see Algorithm~\ref{alg:acc-def}} \label{line:noisy-step}
    \ELSE[precise step]
        \STATE NextState $\gets$ CurrState $\times a$ \label{line:precise-step}
    \ENDIF
    \STATE \textbf{Return} NextState
\end{ALC@g}
\STATE \textbf{Function} \texttt{Reward}(CurrState, \texttt{K}, $a, \sigma$) 
\begin{ALC@g}
    \STATE NextState $\gets$ \texttt{Transition} (CurrState, \texttt{K}, $a, \sigma$)
    \STATE IR $\gets \texttt{IR}_{A}$(NextState) 
    \COMMENT{use Equation~\ref{eq:path-ir-inte} for accumulated plausible noise or Equation~\ref{eq:plau-ir-inte} for one-off plausible noise}
    \IF{$f$(NextState) == $\check{y}$ and $\text{IR} <1-\tau$} \label{line:if-final}
        \STATE Reward $\gets (1-\text{IR})\times\text{ConstantR}$ \label{line:big-reward}
    \ELSE
        \STATE Reward $\gets$ \texttt{softmax}($f$(NextState))     
    \ENDIF
    \STATE \textbf{Return} Reward
\end{ALC@g}
\STATE MDP $\gets \{\mathcal{S}, \mathcal{A}, \texttt{Transition}, \texttt{Reward}, \gamma\}$
\end{algorithmic}
\end{algorithm}

\section{Generating Robust Sequential Recourse}\label{sec:mdp}

In this section, we present a novel approach to address both the one-off plausible noise (introduced in Section~\ref{sec:plau-noise}) and accumulated plausible noise (proposed in Section~\ref{sec:path-noise}) in algorithmic recourse. 
Motivated by the fact that recourse is a sequence of steps in which state transition satisfies the Markov property (Proposition~\ref{prop:acc-noise-dist}), we model the noisy human implementation of recourse as an MDP. We present our proposed method -- RObust SEquential (ROSE) recourse generator 
-- in Algorithm~\ref{alg:rose} and describe it in detail below. 
Its implementation can be found on GitHub\footnote{\sourcecode}.

\paragraph{State Space $\mathcal{S}$}
Given that a dataset has both numerical (Num) and categorical (Cat) features, the state space of our MDP consists of the Cartesian product of the continuous domains for numerical features ($\mathbb{R}^{|\text{Num}|}$) and the discrete domains for categorical features ($\mathbb{Z}^{|\text{Cat}|}$). 

\paragraph{Action Space $\mathcal{A}$}
Actions transition the environment from one state to another. $A(s) \subseteq \mathcal{A}$ denotes all of the actions an agent can execute at state $s\in\mathcal{S}$. To facilitate action sparsity, we assume that each action $a\in A(s)$ corresponds to a modification of a single feature if this action is implemented faithfully. 

\paragraph{Transition Function}
The transition function outputs the state resulting from taking an action $a$ at state $s$. If we assume that there is no noise in the transition, i.e., each action or feature change is implemented faithfully, then the exact feature change resulting from an action is added to the current state, yielding the next state (Algorithm~\ref{alg:rose} Line~\ref{line:precise-step}). On the other hand, if we assume that an action is executed with noise, then it leads to a random state sampled from a plausible noise distribution surrounding $s\times a$ (Algorithm~\ref{alg:rose} Line~\ref{line:noisy-step}).
The former set-up supports the perturbation setting described in Section~\ref{sec:plau-noise}, where only one-off noise is applied to $\check{x}$; %
the latter case is applicable when addressing accumulated plausible noise outlined in Section~\ref{sec:path-noise}. %

\paragraph{Reward Function}
Given the current state and action (as well as the list of implemented actions in the context of accumulated noise), the reward function returns a reward, taking into account whether the resulting state $s^{\prime}$ belongs to the desired class $\check{y}$ and how robust it is to the (accumulated) plausible noise. If the recourse is valid and its degree of robustness meets a certain threshold $\tau$, e.g., less than 25\% of plausible noise is in the undesired class for $\tau=0.75$, then the reward equals to $1-\text{IR}$ multiplied by a large constant reward ConstantR (Algorithm~\ref{alg:rose} Lines~\ref{line:if-final}--\ref{line:big-reward}). If the desired robustness level has not been reached, only a small reward, which equals the probability of $s^{\prime}$ being classified as $\check{y}$, is returned. The small reward steers the algorithm towards a valid counterfactual and speeds up the computation.

\paragraph{Additional Hyper-parameters}
The MDP contains a discount factor $\gamma$, which must be strictly less than 1~\cite{tim2022intro}. Setting $\gamma$ to be less than 1 implicitly rewards shorter paths. %
Other parameters such as noise variance $\sigma^2$ and robustness threshold $\tau$ can be user-defined or domain-specific -- we discuss their specific choices in Section~\ref{sec:experiment}.

\paragraph{Solving MDP Through Policy-gradient Method}
Once we have formulated the MDP, our goal is to find a policy $\pi: \mathcal{S} \rightarrow A$ that, given a state $s\in\mathcal{S}$, returns the best set of actions $A$ to take and gradually guides the agent from $\mathring{x}$ to $\check{x}$. We use existing policy-based deep reinforcement learning methods to approximate the optimal policy $\pi$. Compared with value-based methods, policy-based approaches have better performance when the state space $\mathcal{S}$ or the action space $\mathcal{A}$ are large~\cite{tim2022intro}. Furthermore, policy-gradient methods outperform policy-iteration methods when $\mathcal{S}$ or $\mathcal{A}$ is continuous~\cite{tim2022intro}, which is the case when modelling numerical features. We use a state-of-the-art policy-gradient method, Proximal Policy Optimisation (PPO), with Generalised Advantage Estimate (GAE) to learn the policy~\cite{schulman2017proximal}.

\section{Experimental Evaluation}\label{sec:experiment}

We evaluate ROSE against seven baseline methods on three real-world datasets, and carefully study the effect of various hyper-parameter settings on our results. %

\subsection{Explainers}\label{sec:models}

\paragraph{ROSE}

We set $\gamma=0.99$, which implicitly encourages shorter recourse. We further set \mbox{ConstantR} to 100 so that when a valid counterfactual is reached and the robustness level is satisfied, the agent receives a large reward. %
We test two variants of ROSE. ROSE-one applies one-off plausible noise to $\check{x}$ and follows the precise implementation of actions in $A$. ROSE-mul follows noisy transition between states and relies on accumulated (over multiple steps) plausible noise. 

\paragraph{Baselines}

We first compare ROSE against four non-robust baselines: \citet{wachter2017counterfactual} (Wachter), a gradient-based approach; Growing Sphere (GrSp), which is based on a random search algorithm~\cite{Laugel2018}; DiCE, which generates a diverse set of counterfactual explanations through gradient descent~\cite{mothilal2020explaining}; 
and \textsc{FastAR}, which generates sequential recourse through policy-based reinforcement learning~\cite{verma2022amortized}. The first three methods assume an instantaneous, one-step implementation of the counterfactual explanation, whereas the last one accounts for the order of actions. 
In addition, we evaluate three methods that are designed to generate recourse that is robust to perturbations affecting $\check{x}$: CoGS is an evolutionary algorithm that distinguishes perturbations affecting features that are included in the counterfactuals from the ones that remain unchanged~\cite{virgolin2023robustness}; PROBE uses gradient descent optimisation to generate robust recourse in the presence of noisy human implementation~\cite{pawelczyk2023probabilistic}; and CROCO 
leverages a new estimator of soft recourse invalidation rate that provides a theoretical guarantee on the true recourse robustness~\cite{guyomard2023generating}.

We also experimented with ARAR~\cite{dominguez2022adversarial}, however it is only compatible with linear classifiers such as logistic regression; its validity is as low as 0.02 when the classifier is a neural network. With logistic regression, the length of ARAR recourse is up to ten times worse than that of ROSE, making it impractical for users to follow. \citet{pawelczyk2023probabilistic} reported similar findings about ARAR. Therefore, we do not report its performance in our results. 
All methods use $\ell1$-norm as the distance function during optimisation.

\subsection{Datasets and Classifiers}\label{sec:data-n-clf}
We consider three real-world tabular datasets: German Credit~\cite{german_credit_data}, Adult Income~\cite{becker1996adult} and COMPAS~\cite{larson2016compas}. In each dataset, the desirable class is assigned label $1$, and the undesirable class label $0$. Following best practice, %
all the datasets are normalised so that $x\in[0,1]^d$~\cite{pawelczyk2023probabilistic}.
For prediction models, we train a neural network with a 50-neuron hidden layer and ReLU activation. Every recourse is generated with respect to these classifiers.
We use the learning rate of 0.002, batch size of 50, and 100 training epochs for all the datasets. 
We split the datasets into 80\%--20\% parts used respectively for training and evaluation. The specific configurations of classifiers are provided in Appendix~\ref{app:imp-detail}. 
Given that the German Credit dataset is small, we generate recourse for all the data instances that are predicted with the undesirable class -- hereafter referred to as \emph{negative instances} -- yielding 257 points. On the other hand, the Adult dataset is large, making it computationally expensive for some baselines to output recourse for all the negative instances. To address this issue we randomly choose 200 negative instances from the test set for recourse generation. For COMPAS, we generate recourse for all the 104 negative instances in the test set. %

\begin{table*}[t]
\caption{
ROSE evaluation. We report the average and standard deviation of each metric (except for validity and time) computed across all the chosen negative instances for each dataset (see Section~\ref{sec:data-n-clf} for selection criteria). The best results across \emph{all} the methods are typeset in bold; the best results across the \emph{robust-by-design} methods are underlined. Given the inherent trade-off between robustness and recourse length, our method achieves the highest robustness while effectively managing proximity, i.e., recourse length.
We use DiCE to generate one recourse per negative instance. We set $r=0.35$, $\sigma^2=0.01$ and 30-second search timeout for PROBE. 
In CoGS, we allow features to decrease or increase by 0.1 after normalisation; we use $m=64$ to compute the $\mathcal{K}$-robustness score. We use $K=500$ and $m=0.1$ to allow for the highest confidence level in the IR's upper bound in CROCO. ROSE-one and ROSE-mul use $\sigma^2=0.01$ and $\sigma^2=0.0005$ respectively, and $\tau=0.75$ as the robustness threshold.
}
\centering
\footnotesize%
\begin{tabular}{@{}p{0.1cm}lrrrr>{\color{black}}rrrr@{}}
\toprule
& \multirow{3}{*}{Explainer} & \multicolumn{8}{c}{Metrics} \\ 
    \cmidrule(lr){3-10}
 & & Validity $\uparrow$ & Sparsity $\downarrow$ & Proximity $\downarrow$ & Time (s) $\downarrow$ & Log density $\uparrow$ & Gaussian AIR $\downarrow$ & Plausible AIR $\downarrow$ & Acc AIR $\downarrow$\\ %
 \midrule %
\multirow{7}{*}{\rotatebox[origin=c]{90}{\parbox[c]{3cm}{\centering German Credit}}} 
 & Wachter & \textbf{1.00} & 20.00 $\pm$ 0.00 & 0.46 $\pm$ 0.30 & \textbf{0.020} & -10.31 $\pm$ 0.28 & 0.50 $\pm$ 0.04 & 0.79 $\pm$ 0.16 & 0.75 $\pm$ 0.17 \\ 
 & GrSp & \textbf{1.00} & 16.73 $\pm$ 0.44 & 1.10 $\pm$ 0.78 & \textbf{0.020} & -3.09 $\pm$ 1.46 & 0.43 $\pm$ 0.08 & 0.77 $\pm$ 0.15 & 0.77 $\pm$ 0.18 \\
 & DiCE & \textbf{1.00} & 7.78 $\pm$ 1.40 & 1.12 $\pm$ 0.42 & 0.176 & -8.42 $\pm$ 2.89 & 0.23 $\pm$ 0.16 & 0.26 $\pm$ 0.22 & 0.32 $\pm$ 0.20 \\ 
 & \textsc{FastAR} & 0.93 & \textbf{1.65 $\pm$ 0.70} & \textbf{0.43 $\pm$ 0.34} & 0.040 & 0.34 $\pm$ 1.42 & 0.43 $\pm$ 0.07 & 0.54 $\pm$ 0.11 & 0.51 $\pm$ 0.14 \\
 \cmidrule(lr){3-10}
 & CoGS & \underline{\textbf{1.00}} & 6.98 $\pm$ 1.20 & \underline{0.50 $\pm$ 0.34} & 1.971 &  2.58 $\pm$ 0.42 & 0.39 $\pm$ 0.04 & 0.44 $\pm$ 0.11 & 0.41 $\pm$ 0.14 \\
 & PROBE & \underline{\textbf{1.00}} & 20.00 $\pm$ 0.00 & 1.38 $\pm$ 0.80 & 1.510 & -2.64 $\pm$ 1.20 & 0.24 $\pm$ 0.04 & 0.66 $\pm$ 0.18 & 0.70 $\pm$ 0.27 \\
 & CROCO & 0.49 & 20.00 $\pm$ 0.00 & 1.37 $\pm$ 0.38 & 0.283 & 0.24 $\pm$ 1.43 & \underline{\textbf{0.06 $\pm$ 0.06}} & 0.16 $\pm$ 0.13 & 0.25 $\pm$ 0.13 \\
 & \emph{ROSE-one} & 0.82 & \underline{1.86 $\pm$ 0.84} & 0.54 $\pm$ 0.33 & \underline{0.070} & \underline{\textbf{5.21 $\pm$ 1.98}} & 0.14 $\pm$ 0.05 & 0.22 $\pm$ 0.09 & 0.28 $\pm$ 0.05 \\
 & \emph{ROSE-mul} & 0.71 & 1.95 $\pm$ 0.92 & 0.54 $\pm$ 0.32 & 0.150 & 4.14 $\pm$ 0.58 & \underline{\textbf{0.06 $\pm$ 0.02}} & \underline{\textbf{0.10 $\pm$ 0.04}} & \underline{\textbf{0.23 $\pm$ 0.02}} \\
\midrule%
\multirow{7}{*}{\rotatebox[origin=c]{90}{\parbox[c]{3cm}{\centering Adult Income}}} 
 & Wachter & \textbf{1.00} & 6.00 $\pm$ 0.00 & 0.29 $\pm$ 0.17 & 0.020 & -39.10 $\pm$ 4.29 & 0.50 $\pm$ 0.02 & 0.75 $\pm$ 0.15 & 0.68 $\pm$ 0.18 \\ 
 & GrSp & \textbf{1.00} & 6.46 $\pm$ 0.70 & 0.70 $\pm$ 0.76 & \textbf{0.005} & -14.72 $\pm$ 4.75 & 0.44 $\pm$ 0.08 & 0.68 $\pm$ 0.16 & 0.60 $\pm$ 0.17 \\
 & DiCE & \textbf{1.00} & 5.21 $\pm$ 0.71 & 1.10 $\pm$ 0.51 & 0.125 & -9.57 $\pm$ 5.71 & 0.18 $\pm$ 0.21 & 0.36 $\pm$ 0.30 & 0.45 $\pm$ 0.24 \\
 & \textsc{FastAR} & \textbf{1.00}  & \textbf{1.00 $\pm$ 0.00}     & \textbf{0.09 $\pm$ 0.05}  & 0.010 & 2.21 $\pm$ 1.38 & 0.41 $\pm$ 0.07 & 0.75 $\pm$ 0.14 & 0.51 $\pm$ 0.18 \\
 \cmidrule(lr){3-10}
 & CoGS & \underline{\textbf{1.00}} & 4.94 $\pm$ 0.29 & \underline{0.12 $\pm$ 0.06} & 1.958 & -3.68 $\pm$ 1.54 & 0.41 $\pm$ 0.04 & 0.76 $\pm$ 0.17 & 0.56 $\pm$ 0.27 \\
 & PROBE & 0.99 & 12.98 $\pm$ 0.21 & 12.24 $\pm$ 6.64 & 9.770 & -0.34 $\pm$ 0.50 & 0.35 $\pm$ 0.02 & 0.37 $\pm$ 0.40 & 0.47 $\pm$ 0.19 \\
 & CROCO & 0.99 & 6.00 $\pm$ 0.00 & 0.25 $\pm$ 0.05 & 0.626 & 4.19 $\pm$ 1.95 & 0.09 $\pm$ 0.07 & 0.54 $\pm$ 0.28 & 0.26 $\pm$ 0.25 \\
 & \emph{ROSE-one} & \underline{\textbf{1.00}} & 1.06 $\pm$ 0.24 & 0.18 $\pm$ 0.06 & \underline{0.030} & \underline{\textbf{8.30 $\pm$ 3.19}} & \underline{\textbf{0.05 $\pm$ 0.02}} & \underline{\textbf{0.19 $\pm$ 0.07}} & 0.11 $\pm$ 0.11  \\
 & \emph{ROSE-mul} & \underline{\textbf{1.00}} & \underline{1.02 $\pm$ 0.14} & 0.22 $\pm$ 0.12 & 0.150 & 6.21 $\pm$ 2.35 & 0.08 $\pm$ 0.03 & 0.26 $\pm$ 0.10 & \underline{\textbf{0.04 $\pm$ 0.06}} \\
\midrule%
\multirow{7}{*}{\rotatebox[origin=c]{90}{\parbox[c]{3cm}{\centering COMPAS}}} 
 & Wachter & \textbf{1.00} & 4.00 $\pm$ 0.00 & 0.19 $\pm$ 0.15 & 0.020 & -17.43 $\pm$ 3.85 & 0.47 $\pm$ 0.03 & 0.79 $\pm$ 0.10 & 0.67 $\pm$ 0.16 \\ 
 & GrSp & \textbf{1.00} & 4.05 $\pm$ 0.21 & 0.25 $\pm$ 0.30 & \textbf{0.005} & -6.72 $\pm$ 1.64 & 0.48 $\pm$ 0.03 & 0.75 $\pm$ 0.10 & 0.66 $\pm$ 0.13 \\
 & DiCE & \textbf{1.00} & 3.28 $\pm$ 0.64 & 0.84 $\pm$ 0.51 & 0.063 & -4.49 $\pm$ 0.32 & 0.10 $\pm$ 0.13 & 0.21 $\pm$ 0.16 & 0.31 $\pm$ 0.13 \\
 & \textsc{FastAR} & \textbf{1.00} & \textbf{1.00 $\pm$ 0.00} & \textbf{0.03 $\pm$ 0.01} & 0.050 & -0.61 $\pm$ 1.32 & 0.14 $\pm$ 0.14 & 0.15 $\pm$ 0.14 & 0.05 $\pm$ 0.11 \\
 \cmidrule(lr){3-10}
 & CoGS & \underline{\textbf{1.00}} & 2.88 $\pm$ 0.32 & 0.23 $\pm$ 0.12 & 1.323 & 0.28 $\pm$ 1.44 & 0.27 $\pm$ 0.02 & 0.29 $\pm$ 0.15 & 0.22 $\pm$ 0.15 \\
 & PROBE & \underline{\textbf{1.00}} & 6.99 $\pm$ 0.10 & 0.53 $\pm$ 0.31 & 1.390 & 1.66 $\pm$ 0.83 & 0.33 $\pm$ 0.02 & 0.73 $\pm$ 0.11 & 0.65 $\pm$ 0.16 \\
 & CROCO & \underline{\textbf{1.00}} & 4.00 $\pm$ 0.00 & 0.41 $\pm$ 0.14 & \underline{0.486} & 0.74 $\pm$ 1.43 & 0.10 $\pm$ 0.01 & 0.12 $\pm$ 0.12 & 0.05 $\pm$ 0.07 \\
 & \emph{ROSE-one} & \underline{\textbf{1.00}} & \underline{\textbf{1.00 $\pm$ 0.00}} & \underline{0.04 $\pm$ 0.02} & 0.610 & \underline{\textbf{5.83 $\pm$ 0.57}} & 0.10 $\pm$ 0.08 & 0.11 $\pm$ 0.08 & \underline{\textbf{0.01 $\pm$ 0.03}}  \\
 & \emph{ROSE-mul} & \underline{\textbf{1.00}} & \underline{\textbf{1.00 $\pm$ 0.00}} & \underline{0.04 $\pm$ 0.02} & 0.620 & 4.61 $\pm$ 0.83 & \underline{\textbf{0.09 $\pm$ 0.08}} & \underline{\textbf{0.10 $\pm$ 0.08}} & \underline{\textbf{0.01 $\pm$ 0.03}} \\
\bottomrule
\end{tabular}
\label{tab:acc-noise}
\end{table*}

\subsection{Evaluation Metrics}\label{sec:metrics}%
We employ evaluation metrics that are commonly used in the literature to assess the performance of our explainers~\cite{verma2022amortized,pawelczyk2023probabilistic}. Specifically, we report \textit{validity} -- the fraction of negative instances for which an explainer can successfully generate valid recourse -- and \textit{sparsity} -- the number of features that need to be changed to achieve the recourse. We also measure the \textit{proximity} between $\mathring{x}$ and $\check{x}$ using $\ell1$ distance. 
To empirically evaluate explainers' computational complexity, we report the average \textit{time} taken to generate recourse for a single instance. 
We further report \textit{feasibility} of the generated recourse to quantify its adherence to the geometry of the local data manifold. It is measured using \textit{log density} computed as the logarithmic probability density of the recourse under the target class.

We also check if an explainer is robust to Gaussian and plausible noise, 
which is commonly measured with \textit{Average Invalidation Rate} (AIR), calculated as the probability of obtaining a counterfactual with an undesirable class when small changes (sampled from varying noise distributions) are applied to it~\cite{guyomard2023generating,pawelczyk2023probabilistic}. 
To generate (one-off) Gaussian noise, we randomly sample 1,000 points from a Gaussian distribution $\mathcal{N}(0, \sigma^2\mathbf{I})$ as $\epsilon$ for each $\check{x}$, and then compute IR (as described in Equation~\ref{eq:ir-eq}) across all instances; the results are reported as \textit{Gaussian~AIR}. Similarly, to measure whether an explainer is robust to (one-off) \emph{plausible} noise, we sample 1,000 points from the local data distribution $\mathcal{N}(\check{x}, \sigma^2\mathbf{I}) \cdot K(\mathcal{X})$; this is reported as \textit{Plausible~AIR}. 
In both experiments we set $\sigma^2=0.01$. When evaluating accumulated noise, we use 5\% of the size of the one-off noise (i.e., $\sigma^2=0.0005$) as the noise added to each unit of action. We set the size of the action unit to $u=0.025$ so that the closest recourse $\delta=0.03$ contains at least one action unit. 
For non-sequential methods we assume that the actions (i.e., feature changes) are executed in a random order since these approaches do not output this information. For sequential explainers we follow the action order output by each method. We run the evaluation procedure outlined in Algorithm~\ref{alg:acc-eval} 1,000 times and report its results as \textit{Acc~AIR}. 

\subsection{Implementation Details}

We use off-the-shelf implementation of the PPO with GAE algorithm~\cite{pytorchrl} to solve our MDP problem, and use the OpenAI Gym library~\cite{opaigym} to create the environment for each dataset.
In the PPO algorithm, both actor and critic are approximated by fully connected neural networks with two 64-neuron hidden layers. We follow the hyper-parameter tuning strategy described by \citet{verma2022amortized}; we also follow the best practice of using random data instances from the training set as the starting points for training described by the authors. %
The training time for each dataset is set to half an hour and one hour for ROSE-one and ROSE-mul respectively.
We use CPUs for training since GPUs cannot accelerate the most time-consuming step of our experiments, which is the calculation of IR. Additional implementation details of ROSE are provided in Appendix~\ref{app:imp-detail}. %

For each dataset, we perform a grid search over the training dataset to determine the optimal hyper-parameter settings when applicable. For non-robust methods, we select hyper-parameters for which the method achieves the highest validity. For robust-by-design baseline methods, 
hyper-parameters are chosen based on the lowest Gaussian~AIR given that these methods are built for robustness against Gaussian noise.
For ROSE-one and ROSE-mul, we select hyper-parameters based on the lowest Plausible~AIR and Acc~AIR respectively. 
The details of our hyper-parameter configuration are provided in the caption of Table~\ref{tab:acc-noise}. All the methods, except for the reinforcement learning-based approaches \textsc{FastAR} and ROSE, are implemented using the state-of-the-art package CARLA~\cite{pawelczyk2021carla}.

\subsection{Results}\label{sec:results}

It has been proven theoretically that CE methods face an inherent trade-off 
between their robustness and other desiderata, i.e., increasing robustness of CEs results in a drop of other metrics~\cite{guyomard2023generating,pawelczyk2023probabilistic}. Therefore, it is critical for recourse generators to effectively manage such trade-offs. Our experimental results demonstrate that our method outperforms all the other robustness-oriented approaches across almost all of the metrics and offers the best level of robustness among non-robust approaches, albeit with a slight penalty to the other metrics. 

The performance of all the CE generators is reported in Table~\ref{tab:acc-noise}. For each dataset, the first four methods generate non-robust recourse and the last five approaches generate recourse that is robust to %
different perturbation types. We see that either ROSE-one or ROSE-mul achieves the lowest invalidation rate across all the datasets on Gaussian noise, plausible noise and accumulated plausible noise. Compared with \emph{non-robust} methods, the robustness of ROSE comes at the expense of either worse validity (for German Credit) or increased computation time. On the other hand, the proximity of recourse output by ROSE is comparable with that generated by gradient-based approaches and GrSp for German Credit, and notably smaller for Adult Income and COMPAS. \textsc{FastAR} always produces the shortest recourse with little robustness to any noise; this observation confirms the inherent trade-off between recourse cost (i.e., its proximity) and robustness. Moreover, unlike the gradient-based optimisation methods that modify almost all features, ROSE maintains good sparsity and tends to alter only one or two features across all the datasets. In terms of feasibility, all of the non-robust methods struggle to generate recourse that aligns with the probability density of the underlying data given that they do not explicitly optimise for this property.

Next, we compare ROSE to the \emph{robust-by-design} methods: CoGS, PROBE and CROCO. We always achieve the best sparsity, the shortest computation time (except on COMPAS), the highest log density, and superior robustness for all the perturbation types. ROSE's proximity is the closest on COMPAS and slightly higher than CoGS on German Credit and Adult Income. This suggests that ROSE can more effectively manage the trade-off between the recourse robustness and its cost (i.e., proximity). 
On the other hand, CoGS produces shorter recourse with a high invalidation rate; PROBE has the highest or second highest validity across all the datasets, but its recourse generation time is significantly longer; and CROCO produces relatively proximate recourse in a short time but has high sparsity. While CROCO is more robust to Gaussian noise than CoGS and PROBE, it is less robust than ROSE under every noise distribution.
The results also confirm that guarding against Gaussian noise does not guarantee robustness to plausible noise, even though we used the same $\sigma^2$ value for both types of one-off noise. This is because adapting the Gaussian noise to the local data geometry reshapes the noise distribution, possibly making it markedly different from the normal distribution. In Adult Income, for example, PROBE's AIR to Gaussian and plausible noise is comparable, but in German Credit and COMPAS plausible AIR is more than double that of Gaussian AIR. We find that ROSE-one is robust to plausible noise and at the same time it maintains low level of Gaussian AIR. %

In terms of the accumulated plausible noise, ROSE-mul always achieves the lowest Acc~AIR while at the same time being relatively robust to one-off noise. However, for the German Credit dataset we note that its lower AIR comes at a cost of validity and sparsity. Further, ROSE-mul generally takes more time to generate robust recourse than ROSE-one and the other non-robust explainers. It is also worth noting that robustness to one-off noise does not imply robustness to accumulated noise. Specifically, the Acc~AIR of PROBE is almost as high as that of non-robust methods. This is because PROBE always sacrifices proximity for robustness to one-off noise but disregards the fact that the longer the recourse, the more uncertainty there is, thus it generally produces longer recourse that can accumulate a large amount of noise. 

Furthermore, robust methods are more effective at finding feasible recourse compared to non-robust explainers, even though they do not explicitly incorporate data density into the search process. In general, higher feasibility (log density) comes at an expense of worse proximity, as also observed by \citet{wielopolski2024probabilistically}. Across all datasets, ROSE-one consistently achieves the highest feasibility, followed closely by ROSE-mul.  
We anticipate that when an explainer searches for recourse that is robust to plausible noise, it implicitly explores counterfactuals within dense regions of the data manifold, as the uncertainty around such points is also likely to follow the manifold. 
Additionally, our method ensures recourse actionability; this is achieved by explicitly constraining the action space $\mathcal{A}$ (as explained in Section~\ref{sec:mdp}), which ensures that only actionable features can be modified. By design, ROSE always adheres to these feature value constraints, guaranteeing actionable and feasible recourse.

Since CROCO offers competitive performance across the robustness metrics, it is important to conduct statistical testing to confirm ROSE's superiority. We perform independent two-sample $t$-tests between ROSE-one or ROSE-mul and CROCO. Given that the other methods show markedly worse robustness compared to ROSE, their statistical test results are less relevant. %
For \emph{sparsity}, \emph{proximity} and \emph{log density}, there is no statistical difference between ROSE-one and ROSE-mul, and both ROSE variants significantly outperform CROCO ($p<0.001$) for these three metrics. 
In the case of \emph{Gaussian~AIR}, there is no significant difference between CROCO and ROSE-one or ROSE-mul across the datasets, except for Adult Income where ROSE-one significantly outperforms CROCO ($p<0.001$). 
Regarding \emph{Plausible~AIR}, ROSE-one or ROSE-mul demonstrates statistically significant superiority over CROCO for both German Credit and Adult Income ($p<0.001$). 
Most notably, for \emph{Acc~AIR}, ROSE-mul shows statistically significant improvement over CROCO across all the datasets -- German Credit ($p<0.05$), Adult Income ($p<0.001$), and COMPAS ($p<0.01$). These statistical results highlight that ROSE effectively balances the trade-off between recourse cost (i.e., its length) and robustness, whereas CROCO sacrifices cost to achieve higher robustness against only one-off noise.

In summary, we find that ROSE has the lowest invalidation rate across all the datasets for all the perturbation types among the non-robust recourse methods; it also provides the sparsest and shortest recourse among the robust recourse methods with short compute time.

\begin{figure}[t!]
    \centering
    \begin{subfigure}{0.47\columnwidth}
    \centering
        \includegraphics[height=4.2cm]{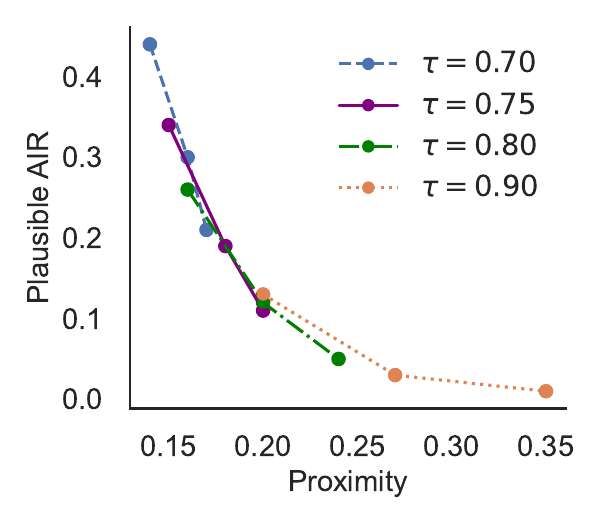}
        \caption{
        In each line, the three dots represent $\sigma^2\in \{0.005, 0.01, 0.015\}$ from left to right.}
        \label{fig:one-plau-params}
    \end{subfigure}
    \hfill
    \begin{subfigure}{0.45\columnwidth}
    \centering
        \includegraphics[height=4.2cm]{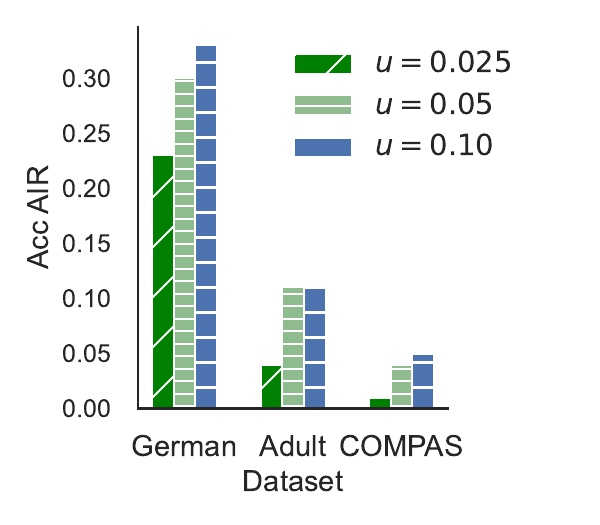}
        \caption{We keep $\sigma^2=0.005$ and $\tau=0.75$ when varying $u$ for each dataset.}
        \label{fig:acc-params}
    \end{subfigure}
    \caption{Panel~(\subref{fig:one-plau-params}) shows the effect of different $\tau$ and $\sigma^2$ when generating robust sequential recourse with ROSE-one for the Adult Income dataset. Panel~(\subref{fig:acc-params}) shows the effect of different $u$ on the robustness of recourse with ROSE-mul.}
    \label{fig:compare-params}
\end{figure}

\subsection{Impact of Hyper-parameters of ROSE}
We further study different hyper-parameters used by ROSE to generate robust recourse. In these experiments, recourse 
is evaluated against the same scale of noise perturbation as described in Section~\ref{sec:metrics}. 

\paragraph{One-off Plausible Noise}
We investigate the effects of $\tau$ and $\sigma^2$ on ROSE-one's ability to generate robust recourse. 
Figure~\ref{fig:one-plau-params} illustrates that %
higher $\tau$ and higher $\sigma^2$ lead to lower AIR, thus better robustness. In these experiments, the validity remains 100\% on the Adult Income dataset. 
Note that we only experiment with $\tau$ up to $0.9$, as setting $\tau$ above this value would further sacrifice validity. 
Our results on Adult Income are representative of the experiments on the other two datasets. 

\paragraph{Accumulated Plausible Noise}
We also study the impacts of $u$, $\tau$ and $\sigma^2$ on ROSE-mul. 
We find that smaller $u$ leads to higher robustness -- see Figure~\ref{fig:acc-params} -- but longer distance. This is because with smaller $u$ each action is assumed to bring smaller feature changes, hence more actions are needed to complete the recourse; as a result, noise is added more frequently. 
$\sigma^2$ and $\tau$ have the same effect on ROSE-one and ROSE-mul. %

\paragraph{Different Classifiers}
We further investigate the performance of ROSE when pairing it with linear classifiers, specifically \emph{logistic regression}. 
We find that the performance of \mbox{ROSE-one} and \mbox{ROSE-mul} does not vary significantly across classifiers, and our approaches consistently outperform all the baselines in terms of robustness and sparsity while at the same time effectively managing recourse cost (i.e., its length) and compute time; for brevity, we omit these results here.

Lastly, we evaluate the robustness of recourse to \emph{varying magnitude of perturbation} while keeping the hyper-parameters used in each method the same as reported in the caption of Table~\ref{tab:acc-noise}. We find that as $\sigma^2$ and $u$ used in evaluation get larger, recourse is perturbed with more noise; consequently, AIR of all the methods increases. Nonetheless, ROSE-one or ROSE-mul still maintain the lowest AIR across all the experiments. The complete set of experimental results can be found in Appendix~\ref{app:add-exp}.  

\section{Conclusion}

Uncertainty and sub-optimal human implementation of algorithmic recourse are often inevitable. When recourse gets longer and more feature changes are required, the process inadvertently becomes more complex and harder for users to implement. Consequently, explainees are less likely to faithfully implement the recommended feature changes. In this paper we argued that the noise affecting recourse -- which captures real-world noisy human actions -- should be plausible and compatible with the underlying data distribution. In addition, we posited that noise accumulates as recourse gets longer, reflecting the increasing difficulty and uncertainty in its implementation. To address these challenges we formulated robust sequential recourse generation as an MDP problem and used policy-based reinforcement learning to generate robust recourse. Our method, called ROSE, accounts for both one-off plausible noise and accumulated plausible noise. It generates a sequence of actions and ensures that even if they are not implemented faithfully -- that is the actual feature changes resulting from user actions are different from the ones prescribed by an explanation -- they will still lead to the desired outcome.

\section*{Declarations}
This research was conducted by the ARC Centre of Excellence for Automated Decision-Making and Society (project number CE200100005), funded by the Australian Government through the Australian Research Council. %
Additional support was provided by the Hasler Foundation (grant number 23082).

\bibliographystyle{ACM-Reference-Format}
\bibliography{reference}

\clearpage\newpage
\appendix

\section{Various Types of Perturbations\label{app:comparison}}
\begin{figure*}[t!]
    \begin{subfigure}{0.4\columnwidth}
        \includegraphics[width=\linewidth]{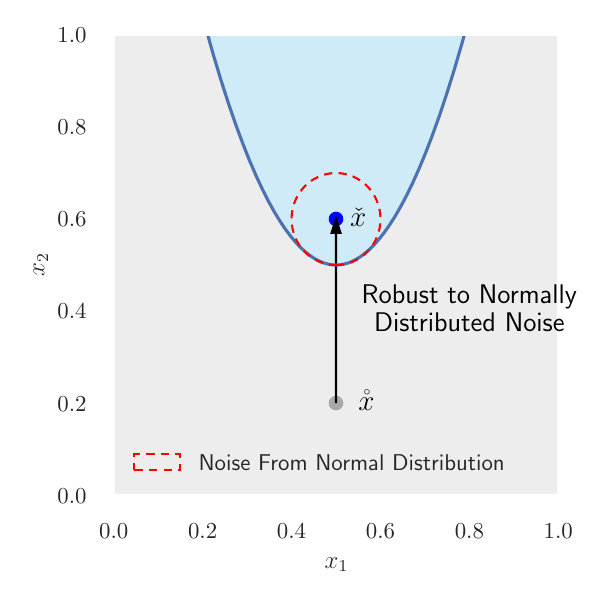}
        \caption{Noise sampled from normal distribution~\citep{pawelczyk2023probabilistic}.}
        \label{fig:normal_noise}
    \end{subfigure}
\hspace{1.5cm}
    \begin{subfigure}{0.4\columnwidth}
        \includegraphics[width=\linewidth]{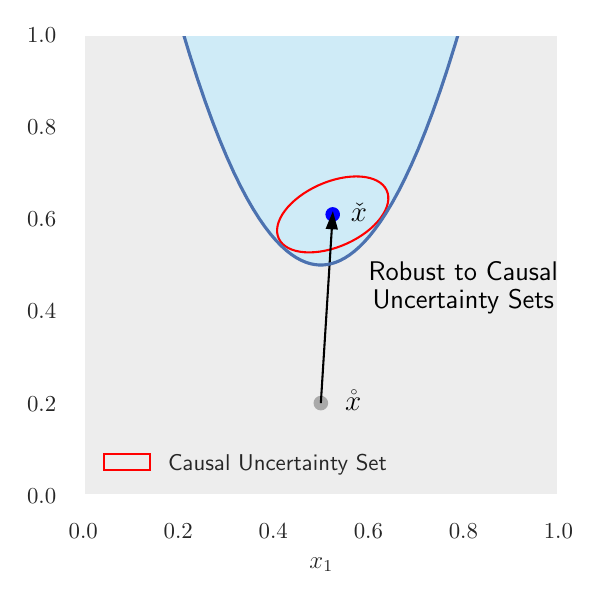}
        \caption{Noise that follows a causal model~\citep{dominguez2022adversarial}.}
        \label{fig:causal_noise}
    \end{subfigure}
    \\
    \begin{subfigure}{0.4\columnwidth}
        \includegraphics[width=\linewidth]{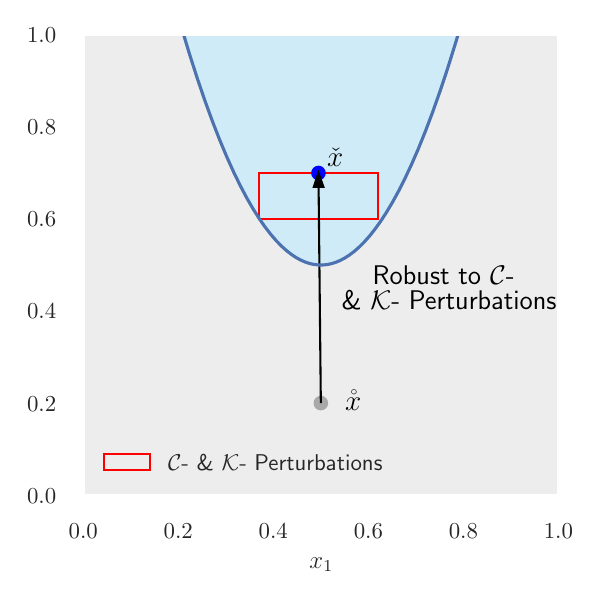}
        \caption{Noise sampled from a manually defined range~\citep{virgolin2023robustness}.}
        \label{fig:boxed_noise}
    \end{subfigure}
\hspace{1.5cm}
    \begin{subfigure}{0.4\columnwidth}
        \includegraphics[width=\linewidth]{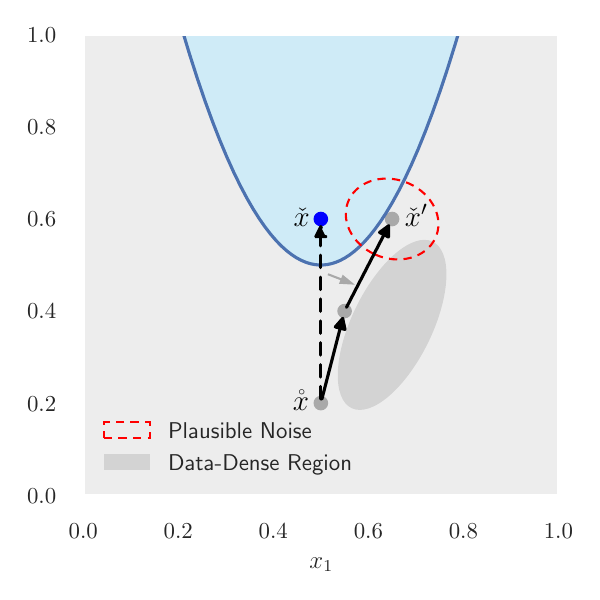}
        \caption{Plausible noise adapted to local data geometry (our work).}
        \label{fig:plausible_noise}
    \end{subfigure}
    \caption{Various ways to represent humans' noisy implementation of algorithmic recourse. In Panels~(\subref{fig:normal_noise}) and~(\subref{fig:plausible_noise}), noise around $\check{x}$ (depicted with dotted lines) follows a probability distribution. In Panels~(\subref{fig:causal_noise}) and~(\subref{fig:boxed_noise}), the shape of the noise distribution (depicted with solid lines) has to be manually defined.
    Additionally, it is assumed that the occurrence of noise is uniformly probably inside the shape outlined by solid red lines regardless of the distance to $\check{x}$.}
    \label{fig:noise_summary}
\end{figure*}

\begin{figure}[t!]
    \centering
    \begin{subfigure}{0.45\columnwidth}
    \centering
        \includegraphics[height=4.5cm]{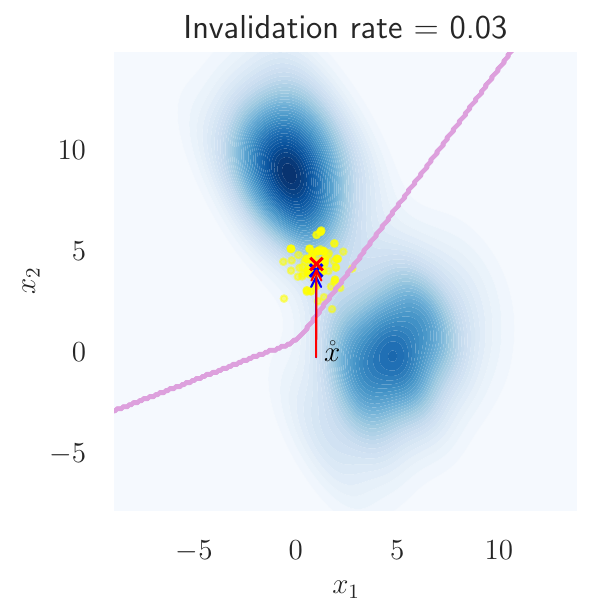}
        \caption{One-step recourse.}
        \label{fig:step-noise-1}
    \end{subfigure}
    \hspace{1em}
    \begin{subfigure}{0.45\columnwidth}
    \centering
        \includegraphics[height=4.5cm]{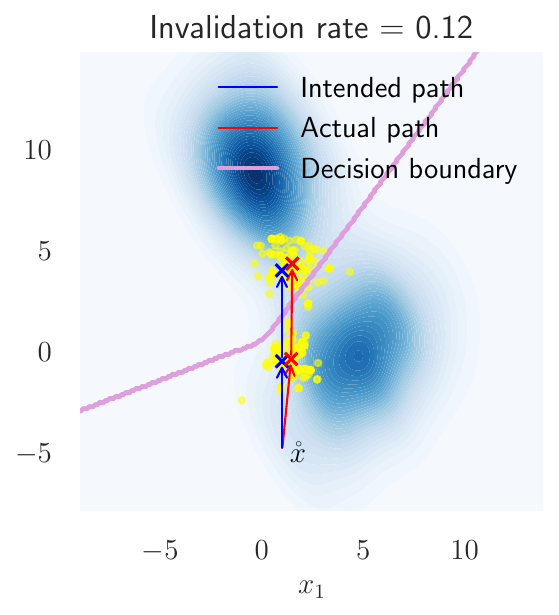}
        \caption{Two-steps recourse.}
        \label{fig:step-noise-2}
    \end{subfigure}
    \caption{Two examples of recourse of varying length generated for the factual instance $\mathring{x}$ and targeting the top-left patch classified with the desired class. The recourse length in Panel~(\subref{fig:step-noise-2}) is double that of the one in Panel~(\subref{fig:step-noise-1}). In Panel~(\subref{fig:step-noise-2}) the noisy state after the first step is determined by sampling from a plausible noise distribution and using the mean of this sample as the new starting point for the subsequent action.}
    \label{fig:step_noise}
\end{figure}

As discussed in Section~\ref{sec:related-work}, current literature on robust recourse uses inconsistent formulations of adversarial perturbations, especially for perturbations applied to individual data instances. In this appendix, we overview the most common representations of noise. We provide a graphical summary of prominent implementations of robust recourse in Figure~\ref{fig:noise_summary}. 

\citet{pawelczyk2023probabilistic} proposed that recourse should be robust to noisy human implementation. To model the noise in human implementation, they perturbed the recourse for a negative data instance by a random variable $\epsilon$ that is drawn from a Gaussian probability distribution (i.e., $\epsilon \sim \mathcal{N}(0, \sigma^2\mathbf{I})$) as shown in Figure~\ref{fig:normal_noise}. However, they assumed the same magnitude of uncertainty in recourse regardless of how difficult it is to implement it. Plausibility was not considered either in their work.

\citet{dominguez2022adversarial} argued that robust recourse should guide \emph{all} instances in the uncertainty set around $\mathring{x}$ to the same desirable predicted outcome. The uncertainty set (i.e., perturbations) includes all of the plausible data points similar to $\mathring{x}$ as shown in Figure~\ref{fig:causal_noise}. To model the plausibility of noise, they explicitly took into account the \emph{linear} causal relationships between features when creating the perturbation set, assuming that the underlying causal model is known or can be well approximated. As such, the distribution of the noise is re-shaped to accommodate the linear causal model. Unlike Gaussian noise in Figure~\ref{fig:normal_noise}, which follows a probability distribution, the uncertainty set in Figure~\ref{fig:causal_noise} assumes that the occurrence of instances within this set is equally probable. 

Similarly, \citet{virgolin2023robustness} accounted for unforeseen circumstances when generating recourse. They manually defined the perturbation ranges for each feature in their experiment dataset based on their domain knowledge. They further broke down the perturbations into two types: perturbations to features that need to be acted upon when implementing the recourse, and perturbations to features that should remain the same when pursuing the recourse. On an abstract level, the perturbation set can be modelled as a hyper-rectangle (Figure~\ref{fig:boxed_noise}) that contains all the possible combinations of feature values within the defined perturbation ranges. Similar to the causal noise, they also assumed that all the instances from this perturbation set are equally probable to draw. 

In this work, we model the sub-optimal recourse implementation and simultaneously consider the plausibility of noisy implementation and accumulated uncertainty. To do so, we associate the magnitude of noise with the number of actions given by recourse and the scale thereof. In the example shown in Figure~\ref{fig:plausible_noise}, the recourse is composed of two steps/actions, each of which may be implemented imperfectly. Thus after taking the first action, the intermediate state could land in a position different from the intended place. When determining the likely positions of any intermediate states, we consider the plausibility of their distribution, i.e., they follow plausible noise distribution. We argue that such plausible noise accumulates as the recourse gets longer, and recourse should be robust to such increasing uncertainty.

We also implement the idea shown in Figure~\ref{fig:plausible_noise} for a synthetic dataset and compute the invalidation rate for plausible noise added to the recourse of varying length. As shown in Figure~\ref{fig:step_noise}, each unit of action is perturbed with the same magnitude of plausible noise. Recourse in Figure~\ref{fig:step-noise-1} is shorter, so plausible noise is only added once. As a result, its invalidation rate is $0.03$. On the other hand, the length of recourse in Figure~\ref{fig:step-noise-2} is double that of the one used in Figure~\ref{fig:step-noise-1}, so plausible noise is added twice -- once per step. As a result, the invalidation rate is $0.12$. Note that the actual path in Figure~\ref{fig:step-noise-2} leans towards the bottom-right cloud of data points, whereas the actual path in Figure~\ref{fig:step-noise-1} is less affected by this cluster of data.

\section{Additional Experiments}\label{app:add-exp}

\subsection{Different Hyper-Parameter Values in ROSE}\label{app:diff-hyper-rose}
In this appendix, we report the performance of ROSE under different hyper-parameter values. Specifically, when generating robust recourse, ROSE uses $\sigma^2$ and $\tau$ to respectively control the magnitude of noise to which recourse should be robust and the degree of target robustness. Once recourse is generated, we evaluate it using noise of the same magnitude.

\paragraph{One-off Plausible Noise}

Figure~\ref{fig:one-plau-params} shows the Average Invalidation Rate (AIR) and proximity when different robustness threshold $\tau$ and variance of plausible noise distribution $\sigma^2$ are used by ROSE-one to generate robust recourse. We can empirically observe that a lower AIR value is achieved at the expense of proximity. In other words, the longer the recourse, the more robust the recourse is to one-off plausible noise. Additionally, higher $\tau$ and $\sigma^2$ lead to lower AIR and better robustness. In these experiments, the validity remains 100\% on the Adult Income dataset. We also observe a similar trade-off in our results on the German Credit and COMPAS datasets. 
Note that we only experiment with $\tau$ up to $0.9$ as setting it to a higher value would further sacrifice validity. Our results also support the argument put forth by \citet{pawelczyk2023probabilistic} that there exists an inherent trade-off between cost (i.e., proximity) and robustness to one-off noise. 

\paragraph{Accumulated Plausible Noise}

When ROSE-mul generates recourse that is robust to the accumulated plausible noise, $\sigma^2$, $u$ and $\tau$ are the hyper-parameters that influence its robustness. In ROSE-mul, $\sigma^2$ is the percentage of the size of one-off noise, which is added to each action unit $u$. Table~\ref{tab:rose-mul-param} shows the performance of ROSE-mul under different $\sigma^2$ and $u$. Bigger $\sigma^2$ indicates that, when generating recourse, each action unit is made robust to bigger noise. %
Bigger $u$ means that fewer actions are used to construct recourse, thus noise is added less frequently. Recourse generated under different hyper-parameters is evaluated with the same scale of noise perturbation. In general, as Acc~AIR decreases, the proximity of recourse degrades. In other words, a higher level of robustness comes at the expense of longer recourse. More robust recourse also requires more compute time. Further, for the German Credit dataset, achieving higher robustness also sacrifices recourse validity to a degree. $\tau$ has the same effect on ROSE-one and ROSE-mul.

\begin{table*}[t!]
\caption{Different values of $\sigma^2$ and $u$ used in ROSE-mul to generate robust recourse. When varying $\sigma^2$, we keep $u=0.025$; when varying $u$, we keep $\sigma^2=5\%$. These results computed for \mbox{ROSE-mul} are complementary to the results shown in Table~\ref{tab:acc-noise}. When evaluating IR for the accumulated noise, we perturb recourse using the same noise specification: $\sigma^2=0.0005$ and $u=0.025$.}
\centering
\small
\begin{tabular}{p{0.3cm}p{1.5cm}rrrrr}
\toprule
& & \multicolumn{5}{c}{Metrics} \\ \cmidrule(lr){3-7}
 & & Validity $\uparrow$ & Sparsity $\downarrow$ & Proximity $\downarrow$ & Time (s) $\downarrow$  & Acc AIR $\downarrow$ \\ \midrule
\multirow{5}{*}{\rotatebox[origin=c]{90}{\parbox[c]{1.8cm}{\centering German Credit}}} 
& $\sigma^2=3\%$ & 0.75 & 1.82 $\pm$ 0.88 & 0.49 $\pm$ 0.32 & 0.171 & 0.26 $\pm$ 0.20 \\
& $\sigma^2= 7\%$ & 0.72 & 1.84 $\pm$ 0.92 & 0.48 $\pm$ 0.34 & 0.173 & 0.21 $\pm$ 0.15 \\
& $\sigma^2 = 10\%$ & 0.70 & 1.83 $\pm$ 0.90 & 0.49 $\pm$ 0.25 & 0.185 & 0.19 $\pm$ 0.18 \\ \cmidrule(lr){2-7}
& $u=0.05$ & 0.78 & 1.71 $\pm$ 0.85 & 0.47 $\pm$ 0.33 & 0.160 & 0.30 $\pm$ 0.19 \\
& $u=0.10$ & 0.81 & 1.71 $\pm$ 0.82 & 0.44 $\pm$ 0.32 & 0.147 & 0.33 $\pm$ 0.28 \\
\midrule
\multirow{5}{*}{\rotatebox[origin=c]{90}{\parbox[c]{1.8cm}{\centering Adult Income}}} 
& $\sigma^2=3\%$ & 1.00 & 1.07 $\pm$ 0.26 & 0.22 $\pm$ 0.09 & 0.179 & 0.08 $\pm$ 0.07 \\
& $\sigma^2= 7\%$ & 1.00 & 1.04 $\pm$ 0.22 & 0.28 $\pm$ 0.10 & 0.196 & 0.05 $\pm$ 0.08 \\
& $\sigma^2 = 10\%$ & 1.00 & 1.03 $\pm$ 0.17 & 0.31 $\pm$ 0.18 & 0.239 & 0.02 $\pm$ 0.05 \\ \cmidrule(lr){2-7}
& $u=0.05$ & 1.00 & 1.02 $\pm$ 0.14 & 0.16 $\pm$ 0.06 & 0.120 & 0.11 $\pm$ 0.09 \\
& $u=0.10$ & 1.00 & 1.01 $\pm$ 0.10 & 0.16 $\pm$ 0.06 & 0.113 & 0.11 $\pm$ 0.10\\\midrule
\multirow{5}{*}{\rotatebox[origin=c]{90}{\parbox[c]{1.8cm}{\centering COMPAS}}} 
& $\sigma^2=3\%$ & 1.00 & 1.00 $\pm$ 0.00 & 0.04 $\pm$ 0.02 & 0.417 & 0.03 $\pm$ 0.08 \\
& $\sigma^2= 7\%$ & 1.00 & 1.00 $\pm$ 0.00 & 0.04 $\pm$ 0.02 & 0.437 & 0.01 $\pm$ 0.03 \\
& $\sigma^2 = 10\%$ & 1.00 & 1.00 $\pm$ 0.00 & 0.04 $\pm$ 0.04 & 0.452 & 0.01 $\pm$ 0.01 \\ \cmidrule(lr){2-7}
& $u=0.05$ & 1.00 & 1.00 $\pm$ 0.00 & 0.04 $\pm$ 0.02 & 0.424 & 0.04 $\pm$ 0.01 \\
& $u=0.10$ & 1.00 & 1.00 $\pm$ 0.00 & 0.04 $\pm$ 0.02 & 0.410 & 0.05 $\pm$ 0.02 \\
\bottomrule
\end{tabular}
\label{tab:rose-mul-param}
\end{table*}

\begin{table*}[hbt!]
\caption{Different values of $\sigma^2$ for one-off plausible noise during evaluation. The configuration of all the methods remains the same as described in Section~\ref{sec:models} and the caption of Table~\ref{tab:acc-noise}. Here, we additionally report Plausible AIR for all the methods under $\sigma^2 \in \{0.015, 0.01, 0.005\}$.}
\centering
\small
\begin{tabular}{p{0.1cm}lrrr}
\toprule
& \multirow{3}{*}{Explainer} & \multicolumn{3}{c}{Plausible AIR} \\
    \cmidrule(lr){3-5}
&  & $\sigma^2 = 0.005$ & $\sigma^2 = 0.01$ & $\sigma^2 = 0.015$ \\  \midrule
\multirow{8}{*}{\rotatebox[origin=c]{90}{\parbox[c]{3cm}{\centering German Credit}}} 
& Wachter & 0.56 $\pm$ 0.11 & 0.79 $\pm$ 0.16 & 0.80 $\pm$ 0.11 \\
& GrSp & 0.52 $\pm$ 0.11 & 0.77 $\pm$ 0.15 & 0.78 $\pm$ 0.11 \\
& DiCE & 0.21 $\pm$ 0.21 & 0.26 $\pm$ 0.22 & 0.32 $\pm$ 0.23 \\
& F\textsc{ast}AR & 0.52 $\pm$ 0.10 & 0.54 $\pm$ 0.11 & 0.55 $\pm$ 0.11 \\
& CoGS & 0.38 $\pm$ 0.09 & 0.44 $\pm$ 0.11 & 0.47 $\pm$ 0.11 \\
& PROBE & 0.47 $\pm$ 0.04 & 0.66 $\pm$ 0.18 & 0.68 $\pm$ 0.10 \\
& CROCO & 0.15 $\pm$ 0.14 & 0.16 $\pm$ 0.13 & 0.25 $\pm$ 0.11\\
& ROSE-one & 0.13 $\pm$ 0.19 & 0.22 $\pm$ 0.09 & 0.28 $\pm$ 0.08 \\
& ROSE-mul & \textbf{0.09 $\pm$ 0.09} & \textbf{0.10 $\pm$ 0.04} & \textbf{0.23 $\pm$ 0.08} \\\midrule
\multirow{8}{*}{\rotatebox[origin=c]{90}{\parbox[c]{2cm}{\centering Adult Income}}} 
& Wachter & 0.73 $\pm$ 0.14 & 0.75 $\pm$ 0.15 & 0.76 $\pm$ 0.15 \\
& GrSp & 0.65 $\pm$ 0.15 & 0.68 $\pm$ 0.16 & 0.69 $\pm$ 0.16 \\
& DiCE & 0.19 $\pm$ 0.26 & 0.36 $\pm$ 0.30 & 0.37 $\pm$ 0.21 \\
& F\textsc{ast}AR & 0.68 $\pm$ 0.15 & 0.75 $\pm$ 0.14 & 0.77 $\pm$ 0.14 \\
& CoGS & 0.70 $\pm$ 0.18 & 0.76 $\pm$ 0.17 & 0.77 $\pm$ 0.16 \\
& PROBE & 0.30 $\pm$ 0.37 & 0.37 $\pm$ 0.40 & 0.36 $\pm$ 0.40 \\
& CROCO & 0.38 $\pm$ 0.16 & 0.54 $\pm$ 0.28 & 0.55 $\pm$ 0.18 \\
& ROSE-one & \textbf{0.05 $\pm$ 0.03} & \textbf{0.19 $\pm$ 0.07} & \textbf{0.30 $\pm$ 0.09} \\
& ROSE-mul & 0.06 $\pm$ 0.05 & 0.26 $\pm$ 0.10 & 0.31 $\pm$ 0.12 \\ \midrule
\multirow{8}{*}{\rotatebox[origin=c]{90}{\parbox[c]{2cm}{\centering COMPAS}}} 
& Wachter & 0.48 $\pm$ 0.18 & 0.79 $\pm$ 0.10 & 0.80 $\pm$ 0.18 \\
& GrSp & 0.48 $\pm$ 0.19 & 0.75 $\pm$ 0.10 & 0.76 $\pm$ 0.06 \\
& DiCE & 0.10 $\pm$ 0.13 & 0.21 $\pm$ 0.16 & 0.24 $\pm$ 0.18 \\
& F\textsc{ast}AR & 0.09 $\pm$ 0.12 & 0.15 $\pm$ 0.14 & 0.18 $\pm$ 0.13 \\
& CoGS & 0.23 $\pm$ 0.12 & 0.29 $\pm$ 0.15 & 0.30 $\pm$ 0.14 \\
& PROBE & 0.51 $\pm$ 0.16 & 0.73 $\pm$ 0.11 & 0.73 $\pm$ 0.17 \\
& CROCO &  0.06 $\pm$ 0.05 & 0.12 $\pm$ 0.12 & 0.17 $\pm$ 0.10 \\
& ROSE-one & \textbf{0.04 $\pm$ 0.05} & \textbf{0.11 $\pm$ 0.08} & \textbf{0.15 $\pm$ 0.09} \\
& ROSE-mul & 0.05 $\pm$ 0.05 & \textbf{0.11 $\pm$ 0.08} & 0.16 $\pm$ 0.09 \\
\bottomrule
\end{tabular}
\label{tab:acc-noise-params-one-sigma}
\end{table*}

\begin{table*}[hbt!]
\caption{Different values of $\sigma^2$ for the accumulated plausible noise with fixed action size unit at $u=0.025$. The configuration of all the methods remains the same as described in Section~\ref{sec:models} and the caption of Table~\ref{tab:acc-noise}, therefore only \mbox{Acc AIR} differs under different $\sigma^2$ in the case of the accumulated plausible noise. In Table~\ref{tab:acc-noise}, $5\%$ of the size of the one-off noise ($\sigma^2=0.0005$) was added to each action unit. Here, we further report \mbox{Acc AIR} of all the methods under $\sigma^2 \in \{0.0003, 0.0005, 0.0007, 0.001\}$, i.e., 3\%, 5\%, 7\% and 10\% of the size of the one-off noise.}
\centering
\small
\begin{tabular}{p{0.1cm}lrrrr}
\toprule
& \multirow{3}{*}{Explainer} & \multicolumn{4}{c}{Acc AIR} \\
    \cmidrule(lr){3-6}
&  & $\sigma^2 = 3\%$ & $\sigma^2 = 5\%$ & $\sigma^2 = 7\%$ & $\sigma^2 = 10\%$ \\  \midrule
\multirow{8}{*}{\rotatebox[origin=c]{90}{\parbox[c]{3cm}{\centering German Credit}}} 
& Wachter & 0.55 $\pm$ 0.10 & 0.75 $\pm$ 0.17 & 0.77 $\pm$ 0.12 & 0.78 $\pm$ 0.11 \\
& GrSp & 0.56 $\pm$ 0.13 & 0.77 $\pm$ 0.18 & 0.78 $\pm$ 0.15 & 0.79 $\pm$ 0.13 \\
& DiCE & 0.27 $\pm$ 0.21 & 0.32 $\pm$ 0.20 & 0.34 $\pm$ 0.21 & 0.36 $\pm$ 0.19 \\
& F\textsc{ast}AR & 0.49 $\pm$ 0.14 & 0.51 $\pm$ 0.14 & 0.52 $\pm$ 0.14 & 0.52 $\pm$ 0.13 \\
& CoGS & 0.37 $\pm$ 0.12 & 0.41 $\pm$ 0.14 & 0.41 $\pm$ 0.11 & 0.44 $\pm$ 0.12 \\
& PROBE & 0.41 $\pm$ 0.16 & 0.70 $\pm$ 0.27 & 0.73 $\pm$ 0.18 & 0.74 $\pm$ 0.16 \\
& CROCO  & 0.23 $\pm$ 0.12 & 0.25 $\pm$ 0.13 & 0.35 $\pm$ 0.07 & 0.38 $\pm$ 0.05 \\
& ROSE-one & 0.21 $\pm$ 0.14 & 0.28 $\pm$ 0.05 & 0.31 $\pm$ 0.15 & 0.34 $\pm$ 0.14 \\
& ROSE-mul & \textbf{0.20 $\pm$ 0.04} & \textbf{0.23 $\pm$ 0.02} & \textbf{0.29 $\pm$ 0.19} & \textbf{0.33 $\pm$ 0.18} \\\midrule
\multirow{8}{*}{\rotatebox[origin=c]{90}{\parbox[c]{3cm}{\centering Adult Income}}} 
& Wachter & 0.65 $\pm$ 0.17 & 0.68 $\pm$ 0.18 & 0.71 $\pm$ 0.18 & 0.72 $\pm$ 0.19 \\
& GrSp & 0.54 $\pm$ 0.17 & 0.60 $\pm$ 0.17 & 0.62 $\pm$ 0.17 & 0.65 $\pm$ 0.16 \\
& DiCE & 0.30 $\pm$ 0.22 & 0.45 $\pm$ 0.24 & 0.48 $\pm$ 0.25 & 0.48 $\pm$ 0.32 \\
& F\textsc{ast}AR & 0.42 $\pm$ 0.16 & 0.51 $\pm$ 0.18 & 0.55 $\pm$ 0.17 & 0.59 $\pm$ 0.17 \\
& CoGS & 0.46 $\pm$ 0.27 & 0.56 $\pm$ 0.27 & 0.61 $\pm$ 0.25 & 0.65 $\pm$ 0.24 \\
& PROBE & 0.45 $\pm$ 0.17 & 0.47 $\pm$ 0.18 & 0.49 $\pm$ 0.16 & 0.51 $\pm$ 0.16 \\
& CROCO & 0.21 $\pm$ 0.12 & 0.26 $\pm$ 0.25 & 0.29 $\pm$ 0.07 & 0.30 $\pm$ 0.18 \\
& ROSE-one & 0.04 $\pm$ 0.07 & 0.11 $\pm$ 0.11 & 0.19 $\pm$ 0.15 & 0.28 $\pm$ 0.17 \\
& ROSE-mul & \textbf{0.01 $\pm$ 0.04} & \textbf{0.04 $\pm$ 0.06} & \textbf{0.09 $\pm$ 0.10} & \textbf{0.14 $\pm$ 0.10} \\ \midrule
\multirow{8}{*}{\rotatebox[origin=c]{90}{\parbox[c]{2cm}{\centering COMPAS}}} 
& Wachter & 0.47 $\pm$ 0.16 & 0.67 $\pm$ 0.16 & 0.67 $\pm$ 0.23 & 0.69 $\pm$ 0.12 \\
& GrSp & 0.45 $\pm$ 0.13 & 0.50 $\pm$ 0.16 & 0.53 $\pm$ 0.17 & 0.54 $\pm$ 0.22 \\
& DiCE & 0.19 $\pm$ 0.12 & 0.31 $\pm$ 0.13 & 0.34 $\pm$ 0.14 & 0.35 $\pm$ 0.13 \\
& F\textsc{ast}AR & 0.03 $\pm$ 0.01 & 0.05 $\pm$ 0.16 & 0.05 $\pm$ 0.11 & 0.05 $\pm$ 0.11 \\
& CoGS & 0.15 $\pm$ 0.14 & 0.22 $\pm$ 0.15 & 0.24 $\pm$ 0.14 & 0.28 $\pm$ 0.15 \\
& PROBE & 0.29 $\pm$ 0.13 & 0.65 $\pm$ 0.16 & 0.68 $\pm$ 0.17 & 0.69 $\pm$ 0.15  \\
& CROCO & 0.04 $\pm$ 0.05 & 0.05 $\pm$ 0.07 & 0.06 $\pm$ 0.04 & 0.08 $\pm$ 0.03 \\
& ROSE-one & \textbf{0.01 $\pm$ 0.02} & \textbf{0.01 $\pm$ 0.03} & \textbf{0.02 $\pm$ 0.04} & \textbf{0.02 $\pm$ 0.05} \\
& ROSE-mul & \textbf{0.01 $\pm$ 0.04} & \textbf{0.01 $\pm$ 0.03} & \textbf{0.02 $\pm$ 0.04} & \textbf{0.02 $\pm$ 0.04} \\
\bottomrule
\end{tabular}
\label{tab:acc-noise-params-sigma}
\end{table*}

\begin{table*}[hbt!]
\caption{Different values of action unit size $u$ for the accumulated plausible noise; $\sigma^2$ is fixed to $\sigma^2=0.0005$.}
\centering
\small
\begin{tabular}{p{0.1cm}lrrrrrr}
\toprule
 & \multirow{3}{*}{Explainer} & \multicolumn{2}{c}{$u=0.025$} & \multicolumn{2}{c}{$u = 0.05$} & \multicolumn{2}{c}{$u = 0.10$} \\
 \cmidrule(lr){3-4}\cmidrule(lr){5-6}\cmidrule(lr){7-8}
&  & \# of steps & Acc AIR & \# of steps & Acc AIR & \# of steps & Acc AIR \\ \midrule
\multirow{8}{*}{\rotatebox[origin=c]{90}{\parbox[c]{3cm}{\centering German Credit}}} 
& Wachter & 18.55 $\pm$ 12.16 & 0.75 $\pm$ 0.17 & 9.27 $\pm$ 6.08 & 0.54 $\pm$ 0.12 & 4.64 $\pm$ 3.04 & 0.51 $\pm$ 0.11 \\
& GrSp & 43.92 $\pm$ 31.19 & 0.77 $\pm$ 0.18 & 21.96 $\pm$ 15.59 & 0.55 $\pm$ 0.13 & 10.98 $\pm$ 7.80 & 0.50 $\pm$ 0.12 \\
& DiCE & 44.83 $\pm$ 16.78 & 0.32 $\pm$ 0.20 & 22.41 $\pm$ 8.39 & 0.26 $\pm$ 0.20 & 22.21 $\pm$ 4.19 & 0.20 $\pm$ 0.18 \\
& F\textsc{ast}AR & 17.22 $\pm$ 13.61 & 0.51 $\pm$ 0.14 & 8.61 $\pm$ 6.81 & 0.49 $\pm$ 0.14 & 4.31 $\pm$ 3.43 & 0.45 $\pm$ 0.14 \\
& CoGS & 19.95 $\pm$ 13.44 & 0.41 $\pm$ 0.14 & 9.98 $\pm$ 6.72 & 0.34 $\pm$ 0.13 & 4.98 $\pm$ 3.36 & 0.26 $\pm$ 0.14 \\
& PROBE & 55.09 $\pm$ 31.91 & 0.70 $\pm$ 0.27 & 27.55 $\pm$ 15.90 & 0.40 $\pm$ 0.18 & 13.78 $\pm$ 7.95 & 0.26 $\pm$ 0.13 \\
& CROCO & 49.21 $\pm$ 17.37 & 0.25 $\pm$ 0.13 & 24.61 $\pm$ 8.69 & 0.23 $\pm$ 0.11 & 12.31 $\pm$ 4.35 & 0.19 $\pm$ 0.16 \\
& ROSE-one & 21.60 $\pm$ 13.21 & 0.28 $\pm$ 0.05 & 10.84 $\pm$ 6.61 & 0.20 $\pm$ 0.13 & 5.42 $\pm$ 3.31 & 0.12 $\pm$ 0.11 \\
& ROSE-mul & 21.60 $\pm$ 12.80 & \textbf{0.23 $\pm$ 0.02} & 10.82 $\pm$ 6.40 & \textbf{0.19 $\pm$ 0.13} & 5.41 $\pm$ 3.21 & \textbf{0.10 $\pm$ 0.11} \\ \midrule
\multirow{8}{*}{\rotatebox[origin=c]{90}{\parbox[c]{2cm}{\centering Adult Income}}} 
& Wachter & 11.48 $\pm$ 6.92 & 0.68 $\pm$ 0.18 & 5.74 $\pm$ 3.46 & 0.63 $\pm$ 0.17 & 2.87 $\pm$ 1.73 & 0.57 $\pm$ 0.16 \\
& GrSp & 28.15 $\pm$ 30.40 & 0.60 $\pm$ 0.17 & 14.10 $\pm$ 15.20 & 0.53 $\pm$ 0.18 & 7.04 $\pm$ 7.60 & 0.45 $\pm$ 0.20 \\
& DiCE & 43.82 $\pm$ 20.40 & 0.45 $\pm$ 0.24 & 21.91 $\pm$ 10.20 & 0.39 $\pm$ 0.21 & 10.96 $\pm$ 5.10 & 0.25 $\pm$ 0.20  \\
& F\textsc{ast}AR & 3.60 $\pm$ 2.01 & 0.51 $\pm$ 0.18 & 1.82 $\pm$ 1.03 & 0.40 $\pm$ 0.18 & 0.93 $\pm$ 0.54 & 0.28 $\pm$ 0.17 \\
& CoGS & 4.61 $\pm$ 2.33 & 0.56 $\pm$ 0.27 & 2.31 $\pm$ 1.16 & 0.43 $\pm$ 0.27 & 1.15 $\pm$ 0.58 & 0.28 $\pm$ 0.24 \\
& PROBE & 489.66 $\pm$ 265.49 & 0.47 $\pm$ 0.18 & 244.83 $\pm$ 132.74 & 0.44 $\pm$ 0.19 & 122.41 $\pm$ 66.37 & 0.40 $\pm$ 0.20 \\
& CROCO & 9.59 $\pm$ 4.87 & 0.26 $\pm$ 0.25 & 4.80 $\pm$ 0.44 & 0.18 $\pm$ 0.17 & 2.41 $\pm$ 0.22 & 0.27 $\pm$ 0.13 \\
& ROSE-one & 7.20 $\pm$ 2.40 & 0.11 $\pm$ 0.11 & 3.60 $\pm$ 1.21 & 0.03 $\pm$ 0.06 & 1.83 $\pm$ 0.63 & \textbf{0.00 $\pm$ 0.01} \\
& ROSE-mul & 8.80 $\pm$ 4.80 & \textbf{0.04 $\pm$ 0.06} & 4.42 $\pm$ 2.42 & \textbf{0.01 $\pm$ 0.04} & 2.21 $\pm$ 1.21 & \textbf{0.00 $\pm$ 0.00} \\ \midrule
\multirow{8}{*}{\rotatebox[origin=c]{90}{\parbox[c]{2cm}{\centering COMPAS}}} 
& Wachter & 7.41 $\pm$ 5.89  & 0.67 $\pm$ 0.16 &  3.70 $\pm$ 2.95 & 0.47 $\pm$ 0.18 & 1.85 $\pm$ 1.47 & 0.46 $\pm$ 0.19  \\
& GrSp & 10.08 $\pm$ 11.93 & 0.50 $\pm$ 0.16 & 5.04 $\pm$ 5.97 & 0.51 $\pm$ 0.16 & 2.52 $\pm$ 2.98 & 0.51 $\pm$ 0.16 \\
& DiCE & 33.76 $\pm$ 20.30 & 0.31 $\pm$ 0.13 & 16.88 $\pm$ 10.15 & 0.26 $\pm$ 0.10 & 8.44 $\pm$ 5.08 & 0.13 $\pm$ 0.10 \\
& F\textsc{ast}AR & 1.20 $\pm$ 0.40 & 0.05 $\pm$ 0.11 & 0.60 $\pm$ 0.20 & 0.03 $\pm$ 0.09 & 0.30 $\pm$ 0.10 & 0.02 $\pm$ 0.08 \\
& CoGS & 9.26 $\pm$ 4.84 & 0.22 $\pm$ 0.15 & 4.63 $\pm$ 2.42 & 0.13 $\pm$ 0.13 & 2.31 $\pm$ 1.21 & 0.05 $\pm$ 0.08 \\
& PROBE & 21.17 $\pm$ 12.49 & 0.65 $\pm$ 0.16 & 10.58 $\pm$ 6.24 & 0.45 $\pm$ 0.16 & 5.29 $\pm$ 3.12 & 0.23 $\pm$ 0.15 \\
& CROCO & 19.74 $\pm$ 8.63 & 0.05 $\pm$ 0.07 & 9.87 $\pm$ 4.32 & 0.03 $\pm$ 0.05 & 4.94 $\pm$ 2.16 & 0.03 $\pm$ 0.08 \\
& ROSE-one & 1.60 $\pm$ 0.80 & \textbf{0.01 $\pm$ 0.03} & 0.80 $\pm$ 0.40 & \textbf{0.01 $\pm$ 0.02} & 0.40 $\pm$ 0.20 & \textbf{0.00 $\pm$ 0.00} \\
& ROSE-mul & 1.60 $\pm$ 0.80 & \textbf{0.01 $\pm$ 0.03} & 0.80 $\pm$ 0.40 & \textbf{0.01 $\pm$ 0.02} & 0.40 $\pm$ 0.20 & 0.01 $\pm$ 0.05 \\
\bottomrule
\end{tabular}
\label{tab:acc-noise-params-u}
\end{table*}

\subsection{Varying Noise Magnitude in Evaluation}\label{app:diff-perturb}

\paragraph{One-off Plausible Noise}

Here, we report the invalidation rate for different sizes of one-off plausible noise when evaluating all the methods; the results are given in Table~\ref{tab:acc-noise-params-one-sigma}. For all the methods across all the datasets, as $\sigma^2$ gets larger, the invalidation rate increases. Notably, ROSE-one or ROSE-mul maintains the lowest invalidation rate across all the experiments.

\paragraph{Accumulated Plausible Noise}

Here, we report the performance of different methods when varying the magnitude of the accumulated plausible noise. Both the unit size of an action -- $u$ -- and the variance of noise added to each action unit -- $\sigma^2$ -- influence the magnitude of accumulated noise used for evaluation. 
Tables~\ref{tab:acc-noise-params-sigma} and~\ref{tab:acc-noise-params-u} report the performance when varying either $\sigma^2$ or $u$, with the other parameter kept fixed. Across all the datasets, Acc AIR of all the methods increases as the value of $\sigma^2$ increases (Table~\ref{tab:acc-noise-params-sigma}); on the other hand, Acc AIR decreases as the value of $u$ increases (Table~\ref{tab:acc-noise-params-u}). ROSE-mul consistently outperforms the other methods in terms of Acc AIR.

\subsection{Performance of Additional Baselines}\label{app:add-baseline}

In this appendix, we report the performance of two additional baselines: ROAR and ARAR. ROAR is designed to be robust against model shifts~\cite{upadhyay2021towards}. We explore whether ROAR is also robust to noisy human implementation of recourse. The other baseline, ARAR, is robust to noisy user input~\cite{dominguez2022adversarial}; however, it is only compatible with logistic regression classifiers, hence its validity is as low as $0.02$ when used with a neural network~\cite{pawelczyk2023probabilistic}. Table~\ref{tab:acc-noise-other} reports the performance of all the baselines that are designed to produce recourse that is robust to varying adversarial events. For a fair comparison, and to showcase that ROSE is model-agnostic, we use here logistic regression for classification.

\begin{table*}[hbt!]
\caption{Comparison of ROSE with two additional baselines -- ROAR and ARAR -- as well as three baselines used earlier in our evaluation -- CoGS, PROBE and CROCO. The underlying predictor is a logistic regression model. For ROAR, we set the learning rate to $0.1$; for ARAR, we set $\epsilon=0.01$. The configuration of the remaining methods is the same as described in the caption of Table~\ref{tab:acc-noise}.}
\centering
\small
\begin{tabular}{p{0.1cm}lrrrrrrr}
\toprule
& \multirow{3}{*}{Explainer} & \multicolumn{7}{c}{Metrics} \\ 
    \cmidrule(lr){3-9}
 &  & Validity $\uparrow$ & Sparsity $\downarrow$ & Proximity $\downarrow$ & Time (s) $\downarrow$ & Gaussian AIR $\downarrow$ & Plausible AIR $\downarrow$ & Acc AIR $\downarrow$ \\ \midrule
\multirow{4}{*}{\rotatebox[origin=c]{90}{\parbox[c]{2cm}{\centering German}}}
 & ROAR & \textbf{1.00} & 20.00 $\pm$ 0.00 & 10.56 $\pm$ 1.95 & \textbf{0.015} & \textbf{0.00 $\pm$ 0.00} & \textbf{0.00 $\pm$ 0.00} & \textbf{0.00 $\pm$ 0.00} \\
 & ARAR & 0.99 & 20.00 $\pm$ 0.00 & 3.83 $\pm$ 0.50 & 0.086 & \textbf{0.00 $\pm$ 0.00} & \textbf{0.00 $\pm$ 0.00} & 0.08 $\pm$ 0.08 \\
 & CoGS & \textbf{1.00} & 6.94 $\pm$ 1.20 & 0.36 $\pm$ 0.23 & 1.622 & 0.39 $\pm$ 0.04 & 0.43 $\pm$ 0.07 & 0.37 $\pm$ 0.11 \\
 & PROBE & \textbf{1.00} & 20.00 $\pm$ 0.00 & 1.16 $\pm$ 0.65 & 1.271 & 0.22 $\pm$ 0.03 & 0.33 $\pm$ 0.06 & 0.38 $\pm$ 0.13 \\
 & CROCO & 0.55 & 20.00 $\pm$ 0.00 & 1.14 $\pm$ 0.40 & 0.198 & 0.11 $\pm$ 0.05 & 0.11 $\pm$0.12 & 0.17 $\pm$ 0.08 \\
 & ROSE-one & 0.65 & \textbf{1.53 $\pm$ 0.79} & 0.33 $\pm$ 0.35 & 0.086 & 0.10 $\pm$ 0.06 & 0.11 $\pm$ 0.07 & 0.14 $\pm$ 0.14 \\
 & ROSE-mul & 0.65 & \textbf{1.53 $\pm$ 0.79} & \textbf{0.32 $\pm$ 0.33} & 0.152 & 0.09 $\pm$ 0.04 & 0.08 $\pm$ 0.06 & 0.19 $\pm$ 0.25 \\
\midrule
\multirow{4}{*}{\rotatebox[origin=c]{90}{\parbox[c]{2cm}{\centering Adult}}} 
 & ROAR & \textbf{1.00} & 6.00 $\pm$ 0.00 & 3.09 $\pm$ 0.72 & 0.041 & \textbf{0.00 $\pm$ 0.00} & \textbf{0.00 $\pm$ 0.00} & 0.06 $\pm$ 0.08 \\
 & ARAR & \textbf{1.00} & 13.00 $\pm$ 0.00 & 2.56 $\pm$ 0.78 & 0.194 & 0.03 $\pm$ 0.01 & 0.30 $\pm$ 0.28 & 0.53 $\pm$ 0.28 \\
 & CoGS & \textbf{1.00} & 4.90 $\pm$ 0.33 & 0.35 $\pm$ 0.19 & 3.864 & 0.38 $\pm$ 0.12 & 0.79 $\pm$ 0.24 & 0.67 $\pm$ 0.24 \\
 & PROBE & \textbf{1.00} & 13.00 $\pm$ 0.00 & 1.64 $\pm$ 1.18 & 2.731 & 0.31 $\pm$ 0.04 & 0.76 $\pm$ 0.19 & 0.75 $\pm$ 0.21 \\
 & CROCO & 0.98 & 6.00 $\pm$ 0.00 & 0.43 $\pm$ 0.07 & 0.634 & 0.08 $\pm$ 0.05 & 0.23 $\pm$ 0.08 & 0.24 $\pm$ 0.13 \\
 & ROSE-one & \textbf{1.00} & \textbf{1.01 $\pm$ 0.10} & \textbf{0.17 $\pm$ 0.04} & \textbf{0.030} & 0.05 $\pm$ 0.02 & 0.17 $\pm$ 0.06 & 0.09 $\pm$ 0.08 \\
 & ROSE-mul & \textbf{1.00} & 1.04 $\pm$ 0.20 & 0.23 $\pm$ 0.10 & 0.186 & 0.05 $\pm$ 0.02 & 0.19 $\pm$ 0.08 & \textbf{0.02 $\pm$ 0.03} \\
\midrule
\multirow{4}{*}{\rotatebox[origin=c]{90}{\parbox[c]{2cm}{\centering COMPAS}}} 
 & ROAR & \textbf{1.00} & 4.00 $\pm$ 4.00 & 1.67 $\pm$ 0.23 & \textbf{0.015} & \textbf{0.00 $\pm$ 0.00} & \textbf{0.00 $\pm$ 0.00} & \textbf{0.00 $\pm$ 0.01} \\
 & ARAR & \textbf{1.00} & 7.00 $\pm$ 0.00 & 1.07 $\pm$ 0.28 & 0.072 & 0.03 $\pm$ 0.01 & 0.07 $\pm$ 0.08 & 0.08 $\pm$ 0.09 \\
 & CoGS & \textbf{1.00} & 2.88 $\pm$ 0.32 & \textbf{0.25 $\pm$ 0.14} & 1.279 & 0.26 $\pm$ 0.02 & 0.27 $\pm$ 0.12 & 0.19 $\pm$ 0.13 \\
 & PROBE & \textbf{1.00} & 7.00 $\pm$ 0.00 & 0.61 $\pm$ 0.36 & 1.231 & 0.34 $\pm$ 0.02 & 0.32 $\pm$ 0.16 & 0.27 $\pm$ 0.15 \\
 & CROCO & \textbf{1.00} & 5.00 $\pm$ 0.00 & 0.54 $\pm$ 0.17 & 0.118 & 0.13 $\pm$ 0.02 & 0.25 $\pm$ 0.17 & 0.26 $\pm$ 0.11 \\
 & ROSE-one & \textbf{1.00} & \textbf{1.00 $\pm$ 0.00} & 0.30 $\pm$ 0.05 & 0.156 & 0.11 $\pm$ 0.07 & 0.21 $\pm$ 0.05 & 0.24 $\pm$ 0.09  \\
 & ROSE-mul & \textbf{1.00} & \textbf{1.00 $\pm$ 0.00} & 0.48 $\pm$ 0.13 & 3.113 & 0.09 $\pm$ 0.04 & 0.21 $\pm$ 0.05 & 0.05 $\pm$ 0.10\\
\bottomrule
\end{tabular}
\label{tab:acc-noise-other}
\end{table*}

\subsection{Detailed Comparison with ROAR and ARAR}\label{app:roar-diss}

In this appendix, we further discuss the performance of ROAR and ARAR reported in Table~\ref{tab:acc-noise-other}. When the underlying classifier is a linear model, both ROAR and ARAR have good performance in finding valid recourse (i.e., validity is at or near to 100\%) in a short time. For ROAR, the invalidation rate for one-off noise and accumulated noise is zero or close to zero across all the datasets. ARAR can maintain a low invalidation rate on the COMPAS and German Credit datasets; however, it achieving high robustness comes at the expense of poor proximity and sparsity. Specifically, recourse produced by ARAR necessitates changing every single feature across all the datasets; ROAR provides recourse that changes at least half of the feature set. In reality, poor sparsity entails high complexity and more implementation effort, which is detrimental to the users. %

\begin{figure}[t!]
    \centering
    \includegraphics[width=.5\linewidth]{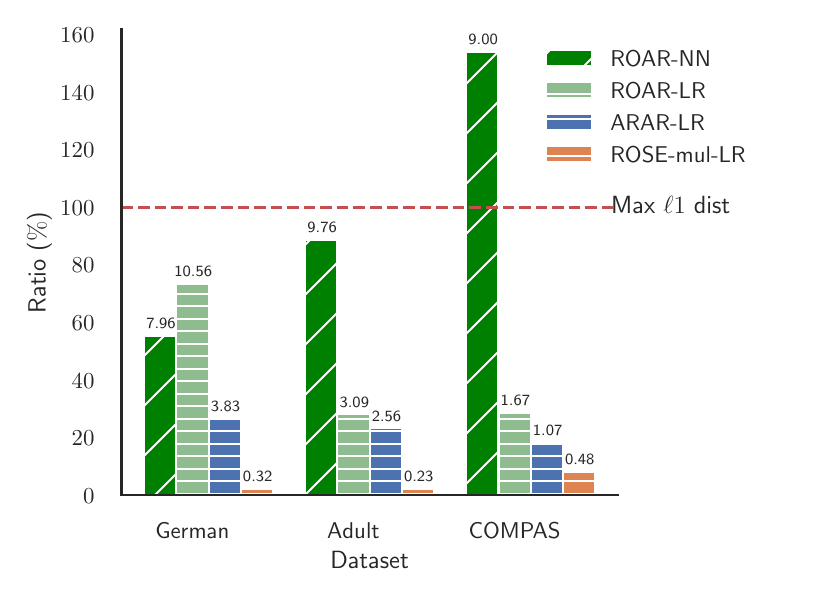}
    \caption{Recourse comparisons for different robust explainers in terms of proximity (i.e., $\ell1$ distance). We report the results of ROAR for both a neural network (NN) and a logistic regression (LR) classifier. ARAR does not support NN, hence the missing results. %
    The distance of recourse produced by ROSE-one and ROSE-mul for NN and LR is similar, hence we only report one result in this figure. The red dotted line indicates the distance between the two points that are farthest apart in each dataset, and the y-axis indicates the ratio of the average recourse distance to the maximum distance between any two data points. The actual $\ell1$ recourse distance is given above each bar.} %
    \label{fig:visual-dist}
\end{figure}

Additionally, ARAR provides recourse with up to 10 times higher distance than our method; for ROAR, this distance can be up to 30 times higher. To better understand the magnitude of such differences, we compare the average distance of recourse provided by ROAR, ARAR and ROSE against the maximum $\ell1$ distance between any two data points found in each dataset. Figure~\ref{fig:visual-dist} shows that for a logistic regression classifier, the $\ell1$ distances in ROAR and ARAR are notably higher than those of ROSE-mul. We further find that for a neural network the average ROAR recourse distance can be longer than the maximum distance between the two data points that are farthest apart in the corresponding dataset. This implies that for ROAR recourse users have to move across all the data points in COMPAS dataset. In summary, even though ROAR and ARAR are robust to noisy human recourse implementation, their explanations are impractical from a user perspective. 

It is also worth noting that ARAR is only compatible with linear classifiers. For non-linear classifiers its validity is as low as 1--2\% across the datasets we considered. ROAR supports non-linear classifiers, but recourse distance is high in such settings. In contrast, our method is model-agnostic and its performance does not vary significantly across different classifiers.

\section{Implementation Details}\label{app:imp-detail}

\subsection{Datasets}

Across all the datasets, we use all the available features for classifier training as well as recourse generation. We acknowledge that some features may be immutable, e.g., sex, and other may be mutable but not actionable, e.g., age; some actionable features may also have actionability constraints. However, the main focus of this work is on robustness, hence we can omit such considerations.
Following the experimental set-up described by \citet{pawelczyk2023probabilistic}, we treat all the features as actionable. In future work we plan to extend our approach to simultaneously address robustness and actionability.

\subsection{Classifiers}%
In Section~\ref{sec:data-n-clf}, we used a neural network as the underlying classifier to determine the negative instances across all the three datasets that we considered. Specifically, our neural network is fully connected with RuLU activation function and a 50-neuron hidden layer. In Appendix~\ref{app:add-baseline} we expanded our analysis by considering a logistic regression as the underlying classifier for all the datasets. The training details for all of our experiments are given in Table~\ref{tab:training-detail}. The performance of the classifiers and the number of negative instances for which recourse was generated are reported in Table~\ref{tab:classifier}.

\begin{table}[t!]
    \caption{Training details of the two classifiers -- Neural Network (NN) and Logistic Regression (LR) -- used in our experiments for the three datasets that we considered.}    
    \centering
    \small
    \begin{tabular}{p{1.8cm}rccc}
    \toprule
       &  & German & Adult & COMPAS \\
         \midrule
         \multirow{2}{*}{Batch size} 
       & LR & 50 & 512 & 50 \\
       & NN & 512 & 512 & 512 \\
       \cmidrule{2-5}
        \multirow{2}{*}{Epochs} 
       & LR & 50 & 50 & 50 \\
       & NN & 50 & 50 & 50 \\
       \cmidrule{2-5}
        \multirow{2}{*}{Learning rate} 
       & LR & 0.001 & 0.001 & 0.001 \\
       & NN & 0.002 & 0.002 & 0.002 \\
       \bottomrule
    \end{tabular}
    \label{tab:training-detail}
\end{table}

\begin{table}[t!]
\caption{Accuracy of our classifiers (top) and the number of negative instances for which recourse was generated (bottom).}%
\centering
\small
\begin{tabular}{p{1.4cm}rccc}
\toprule
   &  & German & Adult & COMPAS \\
     \midrule
     \multirow{2}{*}{Accuracy}
   & LR & 0.78 & 0.84 & 0.86 \\
   & NN & 0.83 & 0.85 & 0.86 \\
       \cmidrule{2-5}
    \multirow{2}{*}{\centering \# of points}
   & LR & 157 & 200 & 97 \\
   & NN & 285 & 200 & 104 \\
   \bottomrule
\end{tabular}
\label{tab:classifier}
\end{table}

\subsection{ROSE Policy-gradient Method Implementation Details}

In this appendix, we provide more details about the policy-gradient method discussed in Section~\ref{sec:mdp}, which we used to solve our MDP problem. We use an off-the-shelf implementation of PPO with GAE algorithm~\cite{pytorchrl}. To leverage this implementation of the PPO algorithm, we needed to create an environment using the open-source toolkit provided by the OpenAI Gym library~\cite{opaigym}. One environment is created for each dataset. In the PPO algorithm, both the actor and critic are approximated with neural networks -- ROSE used a fully connected neural network with two two 64-neuron hidden layers. Set-up of the other hyper-parameters in the PPO algorithm followed the description provided by \citet{verma2022amortized}. For training the PPO algorithm, we also followed the instruction provided by \citet{verma2022amortized}; specifically, we used random data instances from the training set as the starting point as this leads to better RL agent learning~\cite{verma2022amortized}. 

With the MDP set-up described in Section~\ref{sec:mdp}, the total reward gained while training the PPO algorithm to learn the approximated policy converged quite fast. 
For ROSE-one, the one-time training time to learn the policies was 30 minutes for each of the three datasets. For ROSE-mul, the training time was about one hour per dataset. In both cases we used the same number of CPUs. Since the most time-consuming step is calculating IR of the accumulated plausible noise, using GPUs does not accelerate the computation. 

\begin{figure}[t!]
    \centering
    \includegraphics[width=0.8\linewidth]{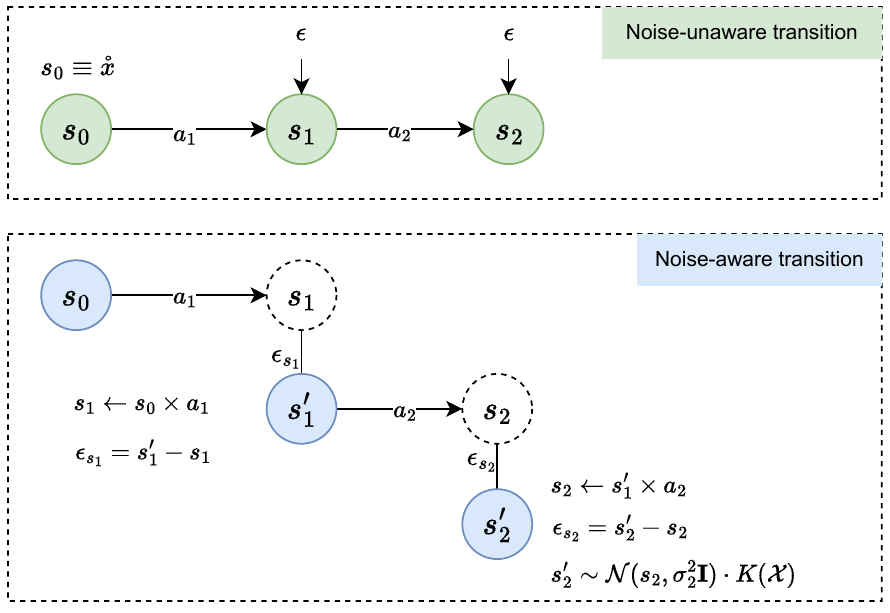}
    \caption{Illustration how different noise can accumulate along recourse %
    with two actions $a_1$ and $a_2$. The green path (top) shows a noise-unaware transition after an action 
    with white Gaussian noise, i.e., $\epsilon \sim \mathcal{N}(0, \sigma^2\cdot\mathbf{I})$, 
    added to each state. The blue path (bottom) shows a noise-aware transition where the subsequent action is applied to the noisy state $s_{1}^{\prime}$. }
    \label{fig:prop-acc-noise-math}
\end{figure}

\section{Proofs}\label{app:proof}

\subsection{Proposition~\ref{prop:acc-noise}}

Equation~\ref{eq:path-noise-dist} shows that $\epsilon_{s_i}$ is sampled based on the distribution $s_i^{\prime} \sim \mathcal{N}(s_i, \sigma_i^2\mathbf{I})\cdot K(\mathcal{X})$, thus the size of $\epsilon_{s_i}$ is proportional to $\sigma_i^2$. Given that $\sigma_i^2 = \frac{||a_i||}{u} \times \sigma^2$, this indicates that
$\epsilon_{s_i} \propto ||a_i||$ %
where $||a_i||$ is the magnitude of feature changes brought by each action. In addition, according to its definition, the accumulated noise is the sum of plausible noise added to each action unit. Namely,
$\mathcal{N}_{(\mathring{x}, \mathcal{A})}(\check{x}) \sim \sum_{i=1}^{k}\epsilon_{s_i}$,
where $k$ is the number of recourse actions (i.e., the number of changed features). 
We can thus conclude that the magnitude of accumulated plausible noise is proportional to the number of changed features and the magnitude of each feature change. 
\qed

\subsection{Proposition~\ref{prop:acc-noise-dist}}

From Definition~\ref{def:path-noise}, we know that $s_i^{\prime}$ is only influenced by $s_{i-1}^{\prime}$ and $a_i$. Once $s_{i-1}^{\prime}$ and $a_i$ are known, the distribution of $s_i^{\prime}$ is \emph{conditionally independent} of all the previous actions $a_{1},\ldots, a_{i-1}$ and intermediate states $s_{1}^\prime,\ldots, s_{i-2}^{\prime}$. Therefore, the noisy transition from one state to the next satisfies the Markov property. 

Figure~\ref{fig:prop-acc-noise-math} provides graphical examples of modelling noisy implementation of recourse with (state-dependent) uniform noise and plausible noise that satisfy the Markov property. 
When we add uniform noise to each action, we can simply ignore the noisy transition between states and sum up all the $\epsilon$ values added to every state as the accumulated noise. On the other hand, when we consider plausible noise, the distribution of which depends on the current state, we need to model the historical state transitions to determine the current state. %
Since $s_2^{\prime}$ only depends on $s_1^{\prime}$ and $a_2$, once these two terms are fixed, the previous state $s_0$ and action $a_1$ do not influence $s_2^{\prime}$. Consequently, we can observe the Markov property in the distribution of plausible noise accumulated along recourse.
\qed

\end{document}